\documentclass{article}

\usepackage{microtype}
\usepackage{graphicx}
\usepackage{subcaption}
\usepackage{booktabs} 
\usepackage{multirow} %
\usepackage{pifont}
\usepackage{hyperref}
\usepackage[table]{xcolor}
\usepackage{enumitem}
\usepackage{tabularx}

\usepackage{algorithm}
\usepackage{algorithmic}
\usepackage{xcolor}
\usepackage{arydshln}

\usepackage[preprint]{icml2026}

\usepackage{amsmath}
\usepackage{amssymb}
\usepackage{mathtools}
\usepackage{amsthm}

\usepackage[capitalize,noabbrev]{cleveref}

\theoremstyle{plain}

\theoremstyle{definition}

\theoremstyle{remark}

\usepackage[textsize=tiny]{todonotes}

\icmltitlerunning{SENSE: Semantic Embedding Navigation with Soft-gated Evaluation}
\usepackage{natbib}

\begin{document}
\twocolumn[
  \icmltitle{SENSE: Semantic Embedding Navigation with Soft-gated Evaluation for Retrieval-based Speculative Decoding}



  \icmlsetsymbol{equal}{*}

  \begin{icmlauthorlist}
    \icmlauthor{Shaowen Chen}{equal,yyy}
    \icmlauthor{Zhicheng Liao}{equal,yyy}
    \icmlauthor{Hongwei Wang}{yyy}
  \end{icmlauthorlist}

  \icmlaffiliation{yyy}{Zhejiang University, Hangzhou, China}

  \icmlcorrespondingauthor{Hongwei Wang}{hongweiwang@intl.zju.edu.cn}

  \icmlkeywords{Machine Learning, ICML}

  \vskip 0.3in
]



\printAffiliationsAndNotice{\icmlEqualContribution}

\begin{abstract}
Speculative Decoding (SD) accelerates Large Language Model (LLM) inference by employing a lightweight draft model to propose candidate tokens, which are verified in parallel by the target model, without compromising generation quality.
While Retrieval-based Speculative Decoding (RSD) is favored for its plug-and-play versatility, its potential is impeded by rigid lexical dependencies, rendering both retrieval and verification brittle to surface-level variations.
To address this, we propose \textit{\textbf{SENSE}} (\textbf{S}emantic \textbf{E}mbedding \textbf{N}avigation with \textbf{S}oft-gated \textbf{E}valuation). 
By anchoring retrieval on the hidden states of the target model, \textit{SENSE} establishes robust semantic alignment, which empowers the Soft-gated Evaluation module to validate semantic equivalence rather than surface forms.
To ensure rigorous benchmarking, we deconstruct existing methods into atomic primitives within a unified framework, facilitating granular, component-level comparison.
Extensive experiments across diverse domains demonstrate that \textit{SENSE} outperforms multiple baselines on the LLaMA and Qwen families, attaining up to $\textbf{4.09}$ mean acceptance length and $\textbf{3.26}\times$ speedup, while preserving generation quality.
Our code will be released upon publication.

\end{abstract}
\section{Introduction}





While large language models (LLMs) leverage parameter scaling to maximize expressivity~\cite{kaplan2020scalinglawsneurallanguage,chowdhery_palm_2022}, their inference efficiency suffers from the sequential nature of auto-regressive decoding~\cite{zeng_glm-130b_2023, Guo_2025}. Among diverse optimization paradigms~\cite{zhou2024surveyefficientinferencelarge}, Speculative Decoding (SD) distinguishes itself through its structural decoupling. By leveraging a lightweight draft model, SD accelerates generation while remaining inherently agnostic to the target model's architecture~\cite{leviathan_fast_2023}.

\begin{figure}[t]
    \centering
    \includegraphics[width=\columnwidth]{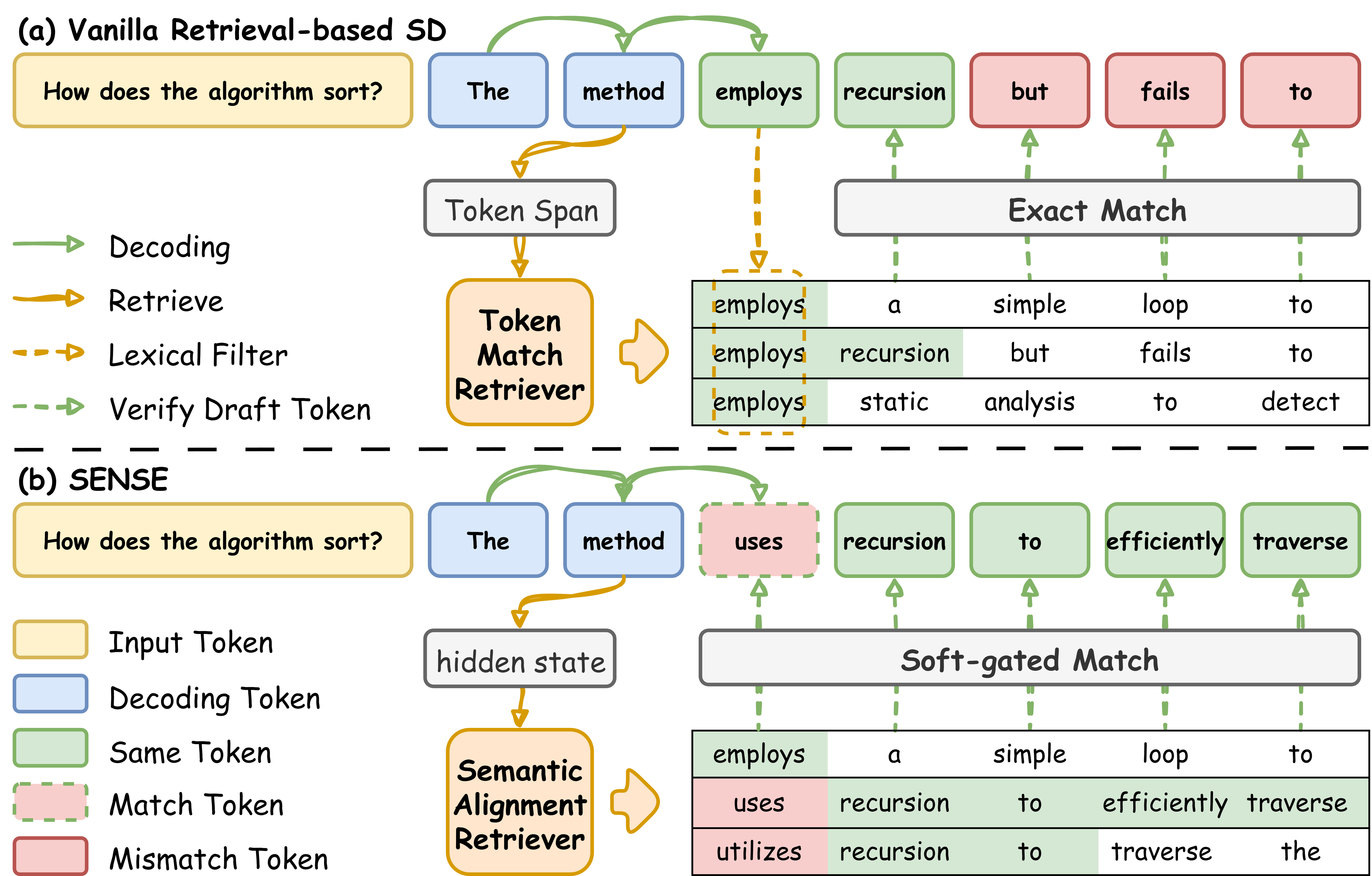}
    \caption{(a) Vanilla RSD retrieves tokens using an exact match rule based on token spans. This approach suffers from lexical limitations, as it strictly requires the retrieved drafts to exactly match the next token "employs". (b) \textit{SENSE} utilizes hidden states to perform semantic alignment retrieval. This allows it to retrieve semantically similar drafts and verify them through a soft-gated match mechanism, effectively overcoming rigid lexical constraints.}
    \label{fig:intro}
\end{figure}

Within the SD landscape, Retrieval-based Speculative Decoding (RSD) stands out for its plug-and-play universality, effective across heterogeneous model architectures and vocabularies without modification~\cite{somasundaram_pld_2024}. However, their flexibility is inherently constrained by the rigidity of exact-match rule. Consequently, many semantically aligned segments are erroneously rejected due to discrete token-level discrepancies. Inspired by lossely SD strategies~\cite{garipov2025autojudgejudgedecodingmanual} that prioritize semantic equivalence over lexical exactness, we posit a critical question: Can the synergy of retrieval-augmented drafting and semantic-aware verification liberate RSD from the limitations of strict lexical matching?

A critical bottleneck persists in the retrieval phase: the rigidity of lexical-based retrieval keys. Mainstream paradigms typically anchor retrieval solely to the target model's immediate token prediction, imposing strict prefix constraints that severely circumscribe the candidate pool~\cite{he_rest_2024, ho_crest_2024}. While recent methodologies like AASD~\cite{wang_alignment-augmented_2025} and SpecLogic~\cite{liu_logitspec_2026} attempt to alleviate this limitation by expanding the search space, they remain confined by the inherent rigidity of token-level exact matching. Consequently, these strategies fail to capture the rich semantic nuances inherent in generation, rendering the retrieval mechanism blind to candidates that are semantically equivalent yet lexically distinct.

Furthermore, the intrinsic nature of retrieved candidates fundamentally constrains verification flexibility. The absence of generative metadata such as specifically probabilistic logits, these candidates are incompatible with advanced distribution alignment heuristics~\cite{noauthor_arc-decode_2025}. Moreover, while retrieval scales to massive candidate pools, the resulting large batch sizes impose severe computational bottlenecks on conventional linear verification strategies~\cite{li_training-free_2025}, thereby negating potential efficiency gains.

In this paper, we introduce \textbf{\textit{SENSE}} (\textbf{S}emantic \textbf{E}mbedding \textbf{N}avigation with \textbf{S}oft-gated \textbf{E}valuation) to overcome these problems. By anchoring on semantic alignment, \textit{SENSE} synergizes an embedding-based retrieval with a complementary soft-gated evaluation.
To overcome lexical rigidity, we introduce \textbf{Semantic Embedding Navigation (SEN)}, which anchors retrieval on the target model's hidden states rather than discrete token predictions. By querying both static datastores and dynamic contexts, \textit{SEN} retrieves candidates aligned with the model's generative intent, even when their surface forms diverge from the greedy prediction.
Complementing the \textit{SEN}, we introduce \textbf{Soft-gated Evaluation (SE)}, an adaptive verification protocol designed to salvage semantically equivalent candidates. 
Driven by entropy-guided logic, \textit{SE} enforces strict exact matching in high-confidence regions while adaptively validating high-uncertainty spans through distributional top-$k$ alignment and parallelized neighborhood fusion to minimize verification overhead. 

Complementing our method, we establish a unified testbed by deconstructing SD into atomic components, enabling the standardized and fair evaluation of diverse baselines. Our code will be released to facilitate reproducible research.



To summarize, our contributions are three-fold. First, we propose \textbf{\textit{SENSE}} (\textbf{S}emantic \textbf{E}mbedding \textbf{N}avigation with \textbf{S}oft-gated \textbf{E}valuation), a training-free framework that overcomes the lexical rigidity of RSD by anchoring retrieval on hidden states and employing entropy-guided verification to accept semantically valid candidates. Second, to ensure fair and reproducible comparison, we implement a systematic orchestration framework that decomposes SD into interchangeable primitives, enabling standardized benchmarking across diverse methodologies. Third, we validate \textit{SENSE} through extensive experiments on LLaMA and Qwen backbones, achieving up to $\textbf{3.26}\times$ speedup with a mean acceptance length of $\textbf{4.09}$ tokens per decoding step.


\section{Related Work}
Retrieval-based speculative decoding accelerates LLM inference through a \textit{retrieve-verify} paradigm: a lightweight retriever fetches candidates from a datastore, followed by parallel verification using the target model~\cite{hu_speculative_2025,ryu_closer_2024}. We structure the related literature along three dimensions: retrieval-based drafting, relaxed verification strategies, and tree-structured parallelism.
\subsection{Retrieval-Based Drafting}
Methods of Retrieval-Based Drafting~\cite{yang_inference_2023,he_rest_2024,hu2024samdecodingspeculativedecoding,ho_crest_2024} fetch candidate continuations from external datastores. Sparse approaches rely on exact n-gram matching, rendering them susceptible to surface-form brittleness; conversely, dense retrieval~\cite{gritta_dresd_2025} mitigates this limitation by leveraging semantic embeddings. Similarly, hybrid strategies~\cite{somasundaram_pld_2024,liu_logitspec_2026,stewart_n-grammys_2024} and heterogeneous architectures~\cite{divilkovskiy_reader_2025} further expand the search scope. However, despite the expanded retrieval space offered by these methods, they remain constrained by \emph{lexical rigidity} during the subsequent filtering and verification phases~\cite{wang_alignment-augmented_2025}. We address this limitation by anchoring retrieval on hidden states and integrating distributional constraints to relax these rigid boundaries.
\subsection{Relaxed Verification Strategies}
Exact-match verification rigidly rejects semantically valid but non-identical drafts~\cite{holtzman2020curiouscaseneuraltext}. Training-based relaxations~\cite{bachmann_judge_2025,garipov2025autojudgejudgedecodingmanual} circumvent strict token matching by learning semantic judgments, whereas training-free approaches~\cite{li_training-free_2025,wang_alignment-augmented_2025,noauthor_arc-decode_2025} exploit entropy metrics or risk bounds. 
Crucially, these methods are constrained by unidimensional criteria along linear paths, causing a cascading loss of valid future tokens upon early failure. We synergize top-$k$ membership and neighborhood-fused error density to robustly verify parallel trajectories.
\subsection{Tree-Structured Parallelism}
Tree-Structured Parallelism~\cite{Miao_2024} amortizes computational costs via prefix sharing. 
Existing methods derive trees from drafters~\cite{cai_medusa_2024,li_eagle_2025,chen_sequoia_2025} or search for high-acceptance structures~\cite{wang_opt-tree_2025,quan_rasd_2025}. 
Traversal verification~\cite{weng_traversal_2025} further preserves subsequences to prevent premature pruning via leaf-to-root acceptance. 
To accommodate dense-retrieved candidates lacking inherent prefix organization, we propose Loose Trie construction via sorted-LCP alignment for efficient compression.
\begin{figure*}[t]
    \centering
    \includegraphics[width=\linewidth]{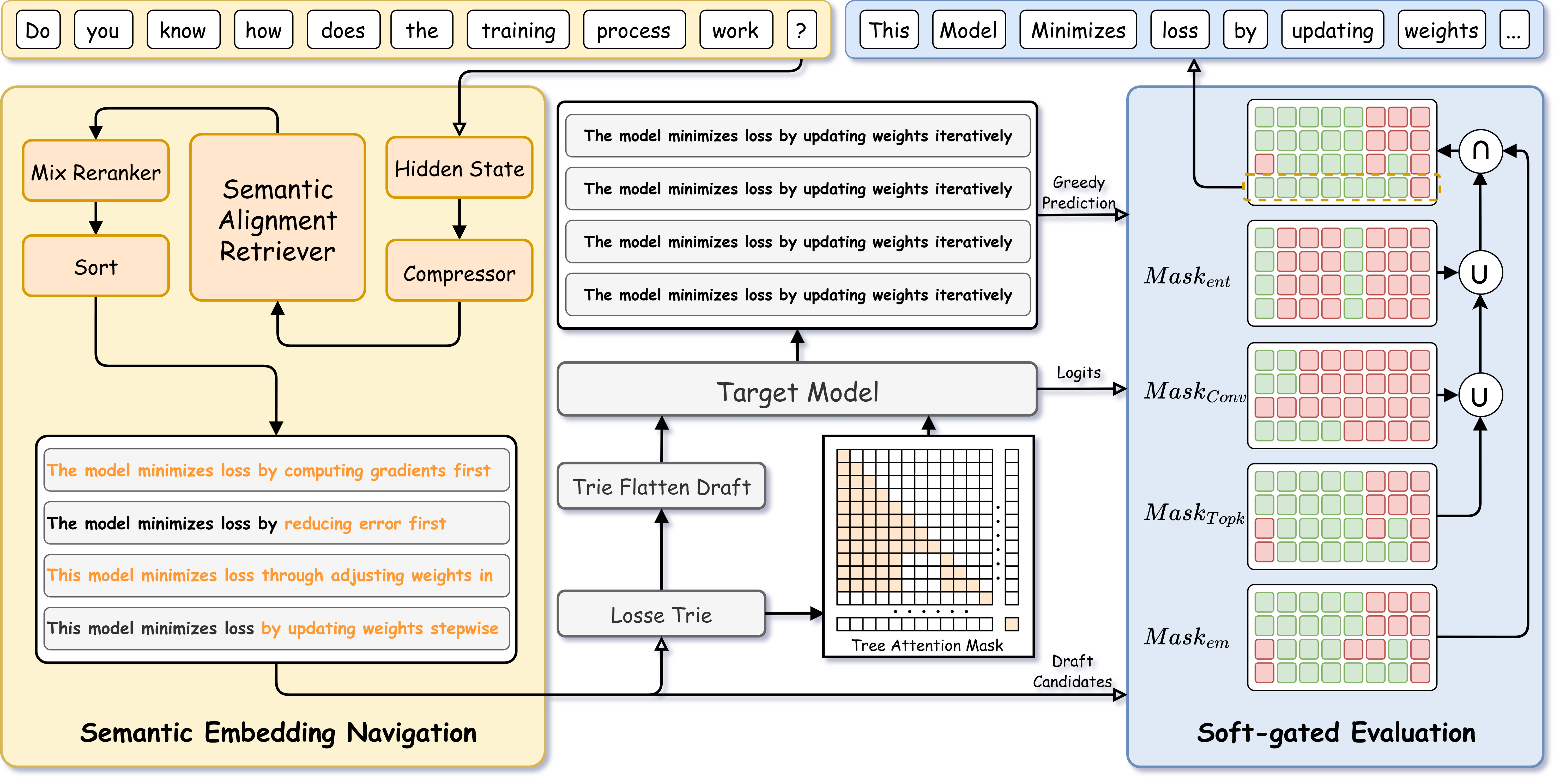} 
    \caption{Overview of the SENSE framework. (Left) The Semantic Embedding Navigation module retrieves semantically aligned drafts based on the input's hidden states. These drafts are flattened into a losse trie structure for efficient verification by the target model. (Right) The Soft-gated Evaluation module filters the verification results using a cascade of masks to robustly determine the final accepted tokens.}
    \label{fig:overview}
\end{figure*}
\section{Method}
Following preliminaries in \cref{sec:2_preliminaries}, we introduce \textit{SENSE} (\cref{fig:framework}), which synergizes Semantic Embedding Navigation (SEN) (\cref{sec:2_drafter}) to augment retrieval diversity and Soft-gated Evaluation (SE) (\cref{sec:2_verifier}) to salvage semantically equivalent yet lexically distinct tokens. Finally, we establish a unified framework to enable fair comparison and reproducible benchmarking across diverse SD paradigms.
\subsection{Preliminaries}
\label{sec:2_preliminaries}
Given the iterative nature of SD, we focus on the operational logic within the generation for a specific token $x_{t}$. 
Let $q_{<t} = [x_1, x_2, \dots, x_{t-1}]$ denote the prefix sequence. 
To generate $x_t$, we process $q_{<t}$ through the target model $LLM$ to obtain the hidden state $h_{q_{<t}}$ and logits $p_{q_{<t}}$:
\begin{equation}
\begin{aligned}
\label{eq:get_hidden_state}
   h_{q_{<t}}, p_{q_{<t}} = LLM(q_{<t})
\end{aligned}
\end{equation}
From these outputs, we derive the greedy anchor $\tilde{x}_{t} = \operatorname*{argmax}_{v \in \mathcal{V}} p_{q_{<t}}[v]$ and the projected query $k_t = F(h_{q_{<t}}, \tilde{x}_{t})$ to identify semantically aligned candidates.
Formally, the corpus $C_{db}$ establishes a mapping from the query embedding space $\mathbb{R}^h$ to text sequences:
\begin{equation}
\begin{aligned}
\label{eq:corpus}
    C_{db} = \{ (k_i, c_i) \mid k_i \in \mathbb{R}^h, c_i \in V^* \}
\end{aligned}
\end{equation}
Building upon this representation, the retriever $R$ identifies the $N$ most relevant sequences  (each with maximum length $M$) from $C_{db}$ according to a similarity metric, constructing the candidate draft set $C_{cand}$:
\begin{equation}
\begin{aligned}
   C_{cand} = R(k_t, C_{db}) = \{c^{1}, c^{2}, \dots, c^{N} \}, |c^{n}| \le M
\end{aligned}
\end{equation}
To verify the candidate set $C_{cand}$, we concatenate each draft with the prefix $q_{<t}$ and execute a batched forward pass through the target model $LLM_{verif}$. This parallel computation yields the probability distributions for all draft positions:
\begin{equation}
\label{eq:verify}
\mathcal{O} = LLM_{verif}(q_{<t}, C_{cand})
\end{equation}
For each draft $c^{n} = [x'_1, \dots, x'_M]$, we can get a sequence of distributions $P^{n} = [p^n_0, p^n_1, \dots, p^n_M]$ from $\mathcal{O}$.
Here, $p^n_0$ corresponds to the position $t$, while $p^n_{i}$ corresponds to position $t+i$ conditioned on the draft prefix.
From these distributions, we derive the greedy verification baseline $\hat{y}^n = [\operatorname{argmax}(p^n_0), \dots, \operatorname{argmax}(p^n_M)]$, which serves as the ground truth for strict matching. 

Finally, the verifier function $V$ determines the acceptance length $\ell^n$ by evaluating the draft $c^n$ against the predicted distributions $P^n$ and the baseline $\hat{y}^n$. The optimal accepted path $d^*$ is selected to maximize this length:
\begin{equation}
\begin{aligned}
\ell^n &= V(c^n, \hat{y}^n, \mathcal{O}) \\
d^* &= \operatorname*{argmax}_{c^n \in C_{cand}} \ell^n, \quad \text{where } 0 \le \ell^n \le M+1
\end{aligned}
\end{equation}
The upper bound $M+1$ accounts for the acceptance of both the anchor token and the full draft. If $\ell^{n} = 0$, the draft is rejected entirely, and the system falls back to the anchor token generated at position $t$.

\subsection{Semantic Embedding Navigation}
\label{sec:2_drafter}
\subsubsection{construct datastore}
To circumvent the lexical rigidity of conventional retrieval, \textit{SEN} anchors candidate retrieval on the target model's hidden states, prioritizing semantic alignment over surface matching.

Building upon \cref{eq:corpus}, we construct a hybrid datastore $C_{db}$ integrating a pre-constructed static corpus ${C}_s$ with a dynamic context buffer ${C}_d$. This unification enables the retrieval of continual suffixes from both historical knowledge and the immediate generation context. 
Specifically, we instantiate the retrieval keys using the model's high-dimensional hidden states.
To mitigate the resulting storage overhead, we employ the projection operator $Z$~\cite{gritta_dresd_2025} to compress these features onto a reduced $v$-dimensional manifold for storage scalability ($v \ll d$):
\begin{equation}
\label{eq:unified_datastore}
\begin{aligned}
    {C}_{db} &= {C}_s \cup {C}_d \\
    &= \{ (Z(h), s) \mid h \in \mathbb{R}^d, s \in V^* \}
\end{aligned}
\end{equation}
With the hybrid datastore in place, we now describe the retrieval procedure that queries it at each decoding step.

\subsubsection{retrieve draft}
At decoding step $t$, we process the prefix $q_{<t}$ via $LLM_{enc}$ to obtain the hidden state $h_{q_{<t}}$ and logits $p_{q_{<t}}$. While $p_{q_{<t}}$ yield the greedy prediction $\tilde{x}_{t}$, $h_{q_{<t}}$ is projected via $Z$ to construct the retrieval query $k_t$. Subsequently, the retriever $R$ executes an approximate nearest neighbor (ANN) search~\cite{douze_faiss_2024} on ${C}_{db}$ to efficiently retrieve the top-$N'$ candidates $D_{ret}$:
\begin{equation}
\label{eq:retrieval_process}
\begin{aligned}
    h_{q_{<t}}, p_{q_{<t}} &= LLM_{enc}(q_{<t}) \\
    k_t &= Z(h_{q_{<t}}) \\
    D_{ret} &= \text{ANN}(k_t, {C}_{db}, N')
\end{aligned}
\end{equation}
The retrieved set $D_{ret} = \{(k_i, s_i)\}_{i=1}^{N'}$ comprises $N'$ key-suffix pairs, where each $s_i \in \mathcal{V}^*$ is a candidate continuation and $k_i$ is its associated embedding. Crucially, candidates are retrieved based on semantic proximity to $k_t$, without requiring their first tokens to match the anchor token $\tilde{x}_t$.

This relaxation introduces candidates with misaligned first tokens. To balance diversity and precision, we propose a composite scoring metric:
\begin{equation} \label{eq:hybrid_scoring}
\begin{gathered}
    S(s_i) = \alpha \cdot \mathbb{I}(s_i[0] = \tilde{x}_{t}) + \beta \cdot \operatorname{sim}(h_{q_{<t}}, k_i) \\
    {D}_{\text{cand}} = \operatorname*{Top-N}_{s_i \in D_{ret}} \left( S(s_i) \right)
\end{gathered}
\end{equation}
The constraint $\alpha \gg \beta$ imposes a quasi-lexicographical ordering, where semantic similarity acts strictly as a secondary discriminator, effectively populating the candidate pool when exact alignments are exhausted.

\subsubsection{Build Losse Trie}
We propose Sorted-LCP Alignment to minimize verification latency~\cite{tan2025specpvimprovingselfspeculativedecoding} by consolidating redundant prefixes. Our approach circumvents the $O(M \times N)$ overhead of canonical Trie construction by employing vectorized adjacent LCP merging. This yields a flattened sequence with mapping $\Gamma$, strictly reducing complexity to $O(\alpha \cdot M)$:
\begin{equation} \label{eq:sorted_lcp}
\begin{gathered}
    \pi = \operatorname*{argsort}_{s_i \in \mathcal{S}_{\text{cand}}} ( s_i ) \\
    l_j = |\operatorname{LCP}(s_{\pi(j)}, s_{\pi(j-1)})| \\
    \mathcal{T}_{\text{flat}} = s_{\pi(1)} \oplus \bigoplus_{j=2}^{N} s_{\pi(j)} [ l_j : ]
\end{gathered}
\end{equation}
Subsequently, we generate a topology-aware mask $M \in \{0, -\infty\}^{L \times L}$ to preserve the causal dependencies within the implicit tree structure where $L = |\mathcal{T}_{\text{flat}}|$~\cite{Miao_2024}. Equipped with the compressed input and $M$, we execute a single forward pass to compute the representations of the flattened sequence with $H \in \mathbb{R}^{L \times d}$ and $P \in \mathbb{R}^{L \times |\mathcal{V}|}$:
\begin{equation}
   (H, P) \in \mathcal{O} = LLM_{verif}(q_{<t}, \mathcal{T}_{flat}; M)
\end{equation}
Finally, we apply the inverse mapping $\Gamma^{-1}$ to restore the flattened representations to the discrete candidate structure. To satisfy the Markov property for the verification of first token $g$, we prepend the predecessor state $h_{t-1}$ and logits $p_{t-1}$ to the reconstructed trajectories, thereby synchronizing the candidate representations with the temporal phase required for parallel verification:
\begin{equation}
\begin{aligned}
    H_{align} &= \left[ h_{t-1} ; \Gamma(H) \right] \in \mathbb{R}^{M \times (N+1) \times d} \\
    P_{align} &= \left[ p_{t-1} ; \Gamma(P) \right] \in \mathbb{R}^{M \times (N+1) \times |\mathcal{V}|}
\end{aligned}
\end{equation}


\subsection{Soft-gated Evaluation}
\label{sec:2_verifier}
$H_{align}$ and $P_{align}$ encapsulate the target predictions.Since \textit{SEN} permits lexical divergence, standard exact-match verification would nullify the utility of semantic retrieval.
To address this, we introduce Soft-gated Evaluation (SE). Grounded in the insight that high entropy and top-$k$ membership signal semantic plausibility~\cite{wang_alignment-augmented_2025,bachmann_judge_2025}, we operationalize relaxed verification via cascaded binary masks using the indicator function $\mathbb{I}(\cdot)$.


\subsubsection{Gating Condition}
We first determine when relaxed verification is warranted. Given the sorted candidate set $S_{\pi}$, we derive discrete token predictions $\mathcal{Y}_{align}$ from $P_{align}$ via greedy decoding, establishing a strict verification baseline:
\begin{equation} 
\label{eq:greedy_decode}
    \mathcal{Y}_{align} = \left[ \operatorname*{argmax}_{v \in \mathcal{V}} P_{align}[i, j, v] \right]_{M \times N}
\end{equation}
Subsequently, the exact-match binary mask $B_{em}$ is computed to isolate tokens satisfying strict verification:
\begin{equation}
    B_{em} = \mathbb{I}(\mathcal{Y}_{align} = S_{\pi}) \in \{0, 1\}^{M \times N}
\end{equation}
To identify candidates eligible for relaxed verification, we employ entropy as a proxy for generative uncertainty, following \cite{wang_alignment-augmented_2025}. High entropy suggests low model confidence, rendering exact matching overly conservative. We compute the entropy matrix $E$ and the gating mask $B_{ent}$ based on a threshold $\gamma$:
\begin{equation}
\begin{gathered}
    E = \left[ -\sum_{v \in \mathcal{V}} P_{align}[i, j, v] \log P_{align}[i, j, v] \right]_{M \times N} \\
    B_{ent} = \mathbb{I}(E > \gamma) \in \{0, 1\}^{M \times N}
\end{gathered}
\end{equation}

\subsubsection{Acceptance Criterion}
Guided by $B_{ent}$, we implement a Top-$k$ criterion~\cite{holtzman2020curiouscaseneuraltext,leviathan_fast_2023} to accommodate plausible semantic alternatives. Recognizing that valid tokens often cluster within the high-probability nucleus despite deviating from greedy predictions, we construct the permissible index set $\mathcal{Y}_{topk}$:
\begin{equation}
\begin{gathered}
    \mathcal{Y}_{topk} = \left[ \operatorname{top-}k_{v \in \mathcal{V}} \big( P_{align}[i, j, v] \big) \right]_{M \times N} \\
    B_{topk} = \mathbb{I}(S_{\pi} \in \mathcal{Y}_{topk}) \in \{0, 1\}^{M \times N}
\end{gathered}
\end{equation}
To further enhance robustness against isolated lexical mismatches, we introduce \textbf{Parallelized Neighborhood Fusion}. Departing from sequential lookahead verification~\cite{li_training-free_2025}, we operationalize the verification logic as a vectorized $1D$ convolution. This mechanism aggregates local exact-matching signals to quantify \textit{local error density}. Specifically, we define the mismatch mask $\bar{B}_{em} = \mathbf{1} - B_{em}$ and convolve it with a unit kernel $\mathbf{k}$ of size $W$:
\begin{equation}
\begin{gathered}
    \mathbf{R} = (\bar{B}_{\text{em}} \oplus \mathbf{1}_{pad}) \ast \mathbf{k}, \quad \text{where } \mathbf{k} = [1]^{W} \\
    B_{\text{con}} = \mathbb{I}(\mathbf{R} \le \tau) \in \{0, 1\}^{M \times N}
\end{gathered}
\end{equation}
Threshold $\tau$ distinguishes acceptable isolated errors (low $\mathbf{R}$) from systematic divergence (high $\mathbf{R}$).

The masks $B_{topk}$ and $B_{con}$ thus provide complementary acceptance criteria: distributional validity via nucleus membership, and structural robustness via local error density.
\subsubsection{Final Decision}
We Synthesize the comprehensive mask $G$ to finalize verification. This formulation integrates the strict verification baseline with a conditional rescue mechanism. Operationally, a token is accepted if it adheres to the exact match criterion ($B_{\text{em}}$); alternatively, conditioned on high uncertainty ($B_{ent}$), it is retained if it exhibits either distributional validity ($B_{topk}$) or local robustness ($B_{con}$):
\begin{equation}
\begin{aligned}
    G &= B_{\text{em}} \lor (B_{ent} \land (B_{topk} \lor B_{con}))
\end{aligned}
\end{equation}
Finally, leveraging the comprehensive mask $G$, we determine the valid prefix length $L_i$ for each draft. This metric is defined by the first rejection index, serving as the basis for selecting the optimal candidate $i^*$:
\begin{equation}
\begin{gathered}
    L_i = \min ( \{ t \mid G[i, t] = 0 \} \cup \{ M+1 \} ) - 1 \\
    i^* = \operatorname*{argmax}_{i \in \{1, \dots, N\}} L_i
\end{gathered}
\end{equation}
Following the verification stage, we determine the input state for the next iteration by indexing into the aligned representations. We define $h_{\text{next}}$ and $p_{\text{next}}$ as the features located at the immediate continuation of the accepted prefix ($L_{i^*} + 1$) within the optimal draft $i^*$:
\begin{equation}
\begin{aligned}
    h_{\text{next}} &= H_{\text{align}}\left[i^*, L_{i^*} + 1\right] \\
    p_{\text{next}} &= P_{\text{align}}\left[i^*, L_{i^*} + 1\right]
\end{aligned}
\end{equation}
The preceding sections detail how SENSE instantiates the drafting and verification phases of speculative decoding. 




\begin{figure}[b]
    \centering
    \includegraphics[width=\columnwidth]{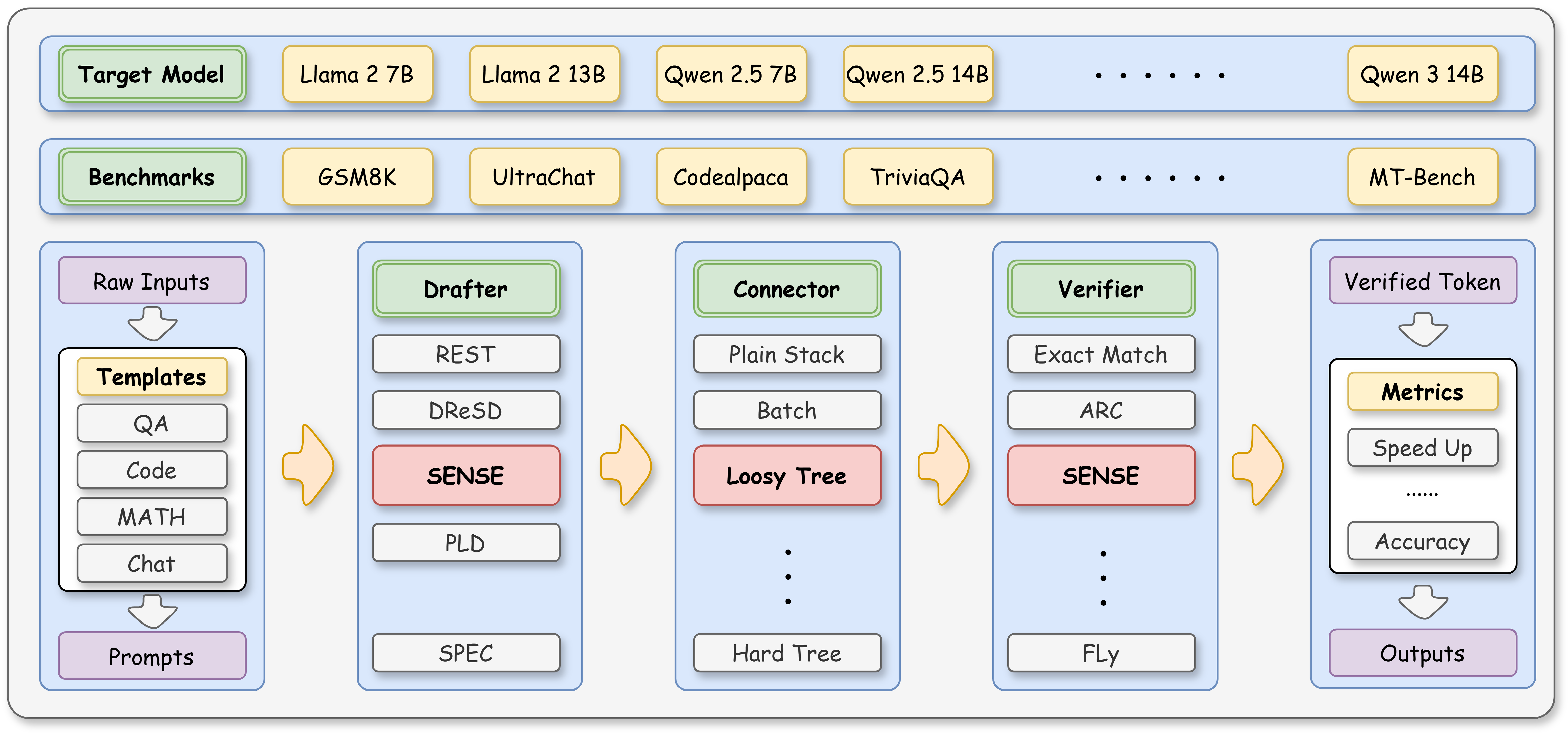} 
    \caption{The atomic framework for SD. We decompose the inference pipeline into modular drafting and verification primitives, orchestrated by a unified connector to enable seamless integration and standardized comparison of diverse methodologies.}
    \label{fig:framework}
\end{figure}

\subsection{Systematic Orchestration Framework}
To facilitate reproducible research and fair comparison across SD, we organize our implementation into three modular stages: \textit{Drafting}, \textit{Connector}, and \textit{Verification} (\cref{fig:framework}).

The \textit{Drafting} stage unifies both retrieval-based methods (REST, DReSD, PLD, and our \textit{SEN}) and generation-based approaches (EAGLE, SpS) under a common interface. The \textit{Connector} transforms heterogeneous draft outputs into a unified tree structure, enabling efficient tree-attention verification regardless of the drafting source. The \textit{Verification} stage supports pluggable acceptance criteria, ranging from strict exact-matching to our proposed soft-gated evaluation.

This modular design serves two practical purposes: (1) enabling controlled ablation studies by swapping individual components, and (2) providing a standardized codebase for future research. Our implementation will be released to facilitate reproducibility.






\begin{table*}[!t]
\renewcommand{\arraystretch}{0.85}
\centering
\setlength{\aboverulesep}{0pt}
\setlength{\belowrulesep}{0pt}
\setlength{\extrarowheight}{1pt}
\scriptsize
\caption{Training-free method comparison reporting Speedup Ratio and Mean Accepted Tokens ($\tau$). Vanilla LLM serves as the 1.0 baseline. \textbf{Bold} and \underline{underlined} text denote the best and second-best results, respectively.}
\label{tab:full_performance}

\resizebox{\textwidth}{!}{
\setlength{\tabcolsep}{3.5pt} 
\begin{tabular}{ll cc cc cc cc cc cc}
\toprule
\multirow{2}{*}{Model} & \multirow{2}{*}{Method} & \multicolumn{2}{c}{GSM8K} & \multicolumn{2}{c}{CodeAlapca} & \multicolumn{2}{c}{UltraChat} &\multicolumn{2}{c}{TriviaQA} & \multicolumn{2}{c}{Mean} \\
\cmidrule(lr){3-4} \cmidrule(lr){5-6} \cmidrule(lr){7-8} \cmidrule(lr){9-10} \cmidrule(lr){11-12}
& & Speedup & $\tau$ & Speedup & $\tau$ & Speedup & $\tau$ & Speedup & $\tau$ & Speedup & $\tau$ \\
\midrule
\multirow{6}{*}{Llama2-7B} 
 & SpS & 1.94 & 2.73 & 1.50 & 2.15 & 1.47 & 2.15 & 1.23 & 1.65 & 1.53 & 2.17  \\
 & PLD & 4.18 & 3.64 & 2.73 & 2.13 & 2.91 & 2.46 & 3.18 & 6.12 & 3.25 & 3.59  \\
 & REST & 2.00 & 1.43 & 2.58 & 1.88 & 2.27 & 1.88 & 2.00 & 1.83 & 2.21 & 1.76  \\
 & DReSD & 1.37 & 1.38 & 1.87 & 2.47 & 1.07 & 1.58 & 1.25 & 1.64 & 1.39 & 1.77  \\
 \rowcolor{gray!25}
 \cellcolor{white} & \textbf{SENSE(OOD)} & \textbf{6.44} & \textbf{6.69} & \textbf{7.88} & \textbf{8.36} & \textbf{4.77} & \textbf{5.50} & \textbf{8.23} & \textbf{8.61} & \textbf{6.83}  & \textbf{7.29} \\
 \rowcolor{gray!25}
 \cellcolor{white} & \textbf{SENSE(ID)} & \underline{5.45} & \underline{5.80} & \underline{6.52} & \underline{6.87} & \underline{3.28} & \underline{3.44} & \underline{7.46} & \underline{7.78} & \underline{5.68} & \underline{5.98} \\

\midrule
\multirow{6}{*}{Llama2-13B} 
 & SpS & 0.73 & 1.84 & 1.42 & 2.25 & 1.31 & 2.00 & 1.57 & 2.44 & 1.26 &  2.13 \\
 & PLD & 1.07 & \textbf{7.37} & 1.54 & \underline{2.79} & 1.54 & 1.95 & \underline{1.70} & \underline{4.75} & \underline{1.46} & \underline{4.22}  \\
 & REST & 0.44 & 1.04 & \underline{1.97} & 2.25 & 1.35 & 1.64 & 0.54 & 1.03 & 1.08 & 1.49 \\
 & DReSD & 0.61 & 1.60 & 0.85 & 2.28 & 1.04 & 2.81 & 1.22 & 3.27 & 0.93 & 2.49  \\
 \rowcolor{gray!25}
 \cellcolor{white} & \textbf{SENSE(OOD)} & \underline{1.12} & 1.20 & 1.95 & 2.08 & \underline{1.64} & \underline{2.84} & 1.09 & 1.16 & 1.46 & 1.82  \\
 \rowcolor{gray!25}
 \cellcolor{white} & \textbf{SENSE(ID)} & \textbf{1.24} & \underline{2.44} & \textbf{3.72} & \textbf{6.83} & \textbf{1.45} & \textbf{2.73} & \textbf{5.36} & \textbf{9.75} & \textbf{2.95} & \textbf{5.44} \\

\midrule
\multirow{6}{*}{Qwen2.5-7B} 
 & SpS & 0.78 & \underline{2.34} & 0.81 & 2.36 & 0.61 & 1.88 & 0.67 & \underline{2.00} & 0.71 & 2.14 &\\
 & PLD & \underline{2.16} & 2.03 & 1.64 & 1.50 & 1.42 & 1.49 & \underline{1.64} & 1.44 & 1.71 & 1.61 \\
 & REST & 1.44 & 1.26 & \underline{2.69} & 2.48 & \underline{1.63} & 1.44 & 1.38 & 1.19 & \underline{1.78} & 1.59 \\
 & DReSD & 1.21 & 1.41 & 2.41 & 1.72 & 1.38 & 1.59 & 1.09 & 1.20 & 1.52 & 1.48  \\
\rowcolor{gray!25}
 \cellcolor{white} & \textbf{SENSE(OOD)} & 1.51 & 1.91 & 2.34 & \underline{3.11} & 1.48 & \underline{2.18} & 1.14 & 1.42 & 1.62 & \underline{2.16} \\
\rowcolor{gray!25}
 \cellcolor{white} & \textbf{SENSE(ID)} & \textbf{2.94} & \textbf{3.44} & \textbf{3.30} & \textbf{3.95} & \textbf{2.59} & \textbf{3.04} & \textbf{2.72} & \textbf{3.63} & \textbf{2.89} & \textbf{3.52} \\

\midrule
\multirow{6}{*}{Qwen2.5-14B} 
 & SpS & 1.09 & \underline{2.26} & 1.44 & 2.31 & 0.88 & \underline{1.82} & 1.01 & \underline{2.08} & 1.10 & \underline{2.11} \\
 & PLD & \underline{2.12} & 2.02 & 1.63 & 1.41 & 1.45 & 1.38 & \underline{1.68} & 0.83 & 1.72 & 1.41  \\
 & REST & 1.38 & 1.23 & \textbf{2.54} & 2.33 & 1.64 & 1.49 & 1.36 & 1.20 & \underline{1.73} & 1.56  \\
 & DReSD & 1.33 & 1.29 & 2.26 & 2.39 & \underline{1.82} & 1.75 & 1.46 & 1.50 & 1.71 & 1.73  \\
 \rowcolor{gray!25}
 \cellcolor{white} & \textbf{SENSE(OOD)} & 1.61 & 1.97 & 2.15 & \underline{2.69}  & 1.62 & 1.74 & 1.25 & 1.53 & 1.66 & 1.98 \\
 \rowcolor{gray!25}
 \cellcolor{white} & \textbf{SENSE(ID)} & \textbf{3.00} & \textbf{3.48} & \underline{2.48} & \textbf{2.90} & \textbf{2.40} & \textbf{2.98} & \textbf{2.47} & \textbf{2.56} & \textbf{2.59} & \textbf{2.96}  \\
\midrule
\multirow{6}{*}{Qwen3-8B} 
 & SpS & 1.34 & \underline{1.98} & 1.61 & 2.02 & 0.68 & 1.70 & 0.72 & \underline{1.75} & 1.09 & \underline{1.86}  \\
 & PLD & 1.41 & 1.81 & 1.30 & 1.35 & 1.26 & 1.37 & 1.25 & 1.38 & 1.30 & 1.48  \\
 & REST & 1.52 & 1.31 & 1.71 & 1.47 & 1.51 & 1.31 & \underline{1.38} & 1.16 & 1.53 & 1.31  \\
 & DReSD & 1.05 & 1.32 & 1.51 & 2.31 & 1.09 & 1.38 & 0.84 & 1.22 & 1.12 & 1.56 \\
 \rowcolor{gray!25}
 \cellcolor{white} & \textbf{SENSE(OOD)} & \underline{1.63} & 1.73 & \underline{1.86} & \underline{2.34} & \underline{1.62} & \underline{1.75} & 1.12 & 1.36 & \underline{1.56} & 1.80 \\
 \rowcolor{gray!25}
 \cellcolor{white} & \textbf{SENSE(ID)} & \textbf{2.48} & \textbf{2.63} & \textbf{2.75} & \textbf{3.02} & \textbf{2.09} & \textbf{2.18} & \textbf{2.22} & \textbf{2.13} & \textbf{2.39} &  \textbf{2.49} \\

\midrule
\multirow{6}{*}{Qwen3-14B} 
 & SpS & 0.58 & \underline{2.27} & 1.05 & \underline{1.99} & 0.60 & \underline{1.79} & 0.52 & \underline{1.92} & 0.69 & \underline{1.99}  \\
 & PLD & 0.91 & 1.71 & \underline{1.53} & 1.39 & 1.05 & 1.31 & 0.82 & 1.37 & 1.08 & 1.44 \\
 & REST & 0.86 & 1.31 & 1.09 & 1.14 & 1.23 & 1.28 & 0.60 & 1.16 & 0.95 & 1.22 \\
 & DReSD & 0.69 & 1.32 & 1.20 & 1.41 & 1.03 & 1.38 & 0.68 & 1.21 & 0.90 & 1.33 \\
 \rowcolor{gray!25}
 \cellcolor{white} & \textbf{SENSE(OOD)} & \underline{0.98} & 1.67 & 1.21 & 1.63 & \underline{1.83} & 1.59 & \underline{1.37} & 1.31 & \underline{1.35} & 1.55 \\
 \rowcolor{gray!25}
 \cellcolor{white} & \textbf{SENSE(ID)} & \textbf{2.73} & \textbf{3.35} & \textbf{3.23} & \textbf{4.03} & \textbf{3.26} & \textbf{6.57} & \textbf{2.89} & \textbf{2.55} & \textbf{3.03} & \textbf{4.13}  \\
\bottomrule
\end{tabular}
}
\vspace{-1em}
\end{table*}

\section{Experiments}
\subsection{Experimental Setup}

\textbf{Benchmarks.} We evaluate across four domains: mathematics (GSM8K~\cite{cobbe_training_2021}), coding (CodeAlpaca~\cite{chaudhary_sahil280114codealpaca_2026}), dialogue (UltraChat~\cite{ding_enhancing_2023}), and question answering (TriviaQA~\cite{joshi_triviaqa_2017}). Critically, we construct both In-Distribution (ID) and Out-Of-Distribution (OOD) datastores: ID datastores contain responses generated by the target LLM itself, while OOD datastores use original ground-truth labels. This distinction enables rigorous assessment of datastore-model alignment effects. Dataset details are in~\cref{append:data}.

\textbf{Models and Baselines.} We evaluate on six target models across two spanning families: Llama-2 (7B/13B)~\cite{touvron_llama_2023} and Qwen (2.5-7B/14B-Instruct~\cite{qwen2.5}, 3-8B/14B~\cite{qwen3technicalreport}). For generative baselines, we pair these with corresponding draft models (MicroLlama~\cite{wang2024microllama}, Qwen2.5-0.5B-Instruct, Qwen3-0.6B). We compare against retrieval-based methods (REST~\cite{he_rest_2024}, DReSD~\cite{gritta_dresd_2025}, PLD~\cite{somasundaram_pld_2024}), generative drafting (SpD~\cite{leviathan_fast_2023}, EAGLE-3~\cite{li_eagle_2025}), and relaxed verification approaches (ARC~\cite{noauthor_arc-decode_2025}, FLY~\cite{li_training-free_2025}). All baselines use their reported optimal configurations, with details in~\cref{append:baselines}.

\textbf{Implementation.} Hidden states are projected via PCA and indexed using FAISS~\cite{douze_faiss_2024} with IVF-PQ. At inference, we retrieve $k{=}3$ candidates with maximum draft length $n{=}10$. Verification uses entropy threshold $\theta_e{=}0.05$, with mismatch window $w{=}6$. All methods use greedy decoding with temperature=0 and FP32 precision. Complete hyperparameters specifications are detailed in~\cref{append:implementation}.

\begin{table}[t]{}
\centering
\caption{Performance comparison with training-based methods. \textbf{Bold} indicates the best performance in each column.}
\scriptsize
\label{tab:training_based_comp}

\resizebox{\columnwidth}{!}{
\setlength{\tabcolsep}{2pt} 
\begin{tabular}{ll cc cc cc}
\toprule
\multirow{2.5}{*}{Model} & \multirow{2.5}{*}{Method} & \multicolumn{2}{c}{GSM8K} & \multicolumn{2}{c}{CodeAlpaca} & \multicolumn{2}{c}{Mean} \\

\cmidrule(lr){3-4} \cmidrule(lr){5-6} \cmidrule(lr){7-8}
& & Speedup & $\tau$ & Speedup & $\tau$ & Speedup & $\tau$ \\
 \midrule
\multirow{2}{*}{Qwen2.5-14B} 
 & EAGLE-2 & 2.19 & 2.43 & 1.49 & \textbf{2.94 }& 1.84 & 2.69 \\
 & \textbf{SENSE(ID)} & \textbf{3.00} & \textbf{3.48} & \textbf{2.48} & 2.90 & \textbf{2.74} &  \textbf{3.19} \\
\midrule
 \multirow{2}{*}{Qwen3-8B} 
 & EAGLE-3 & 1.71 & 2.00 & 1.77 & 2.02 & 1.74 & 2.01 \\
 & \textbf{SENSE(ID)} & \textbf{2.48} & \textbf{2.63} & \textbf{2.75} & \textbf{3.02} & \textbf{2.61} & \textbf{2.83} \\
 \midrule
 \multirow{2}{*}{Qwen3-14B} 
 & EAGLE-3 & 1.65 & 1.87 & 1.51 & 1.78 & 1.58 & 1.83 \\
 & \textbf{SENSE(ID)} & \textbf{2.73} & \textbf{3.35} & \textbf{3.23} & \textbf{4.03} & \textbf{2.98} & \textbf{3.69} \\

\bottomrule
\end{tabular}
}
\end{table}





\begin{figure}[b]
\vspace{-1.5em}
    \centering
    \includegraphics[width=\columnwidth]{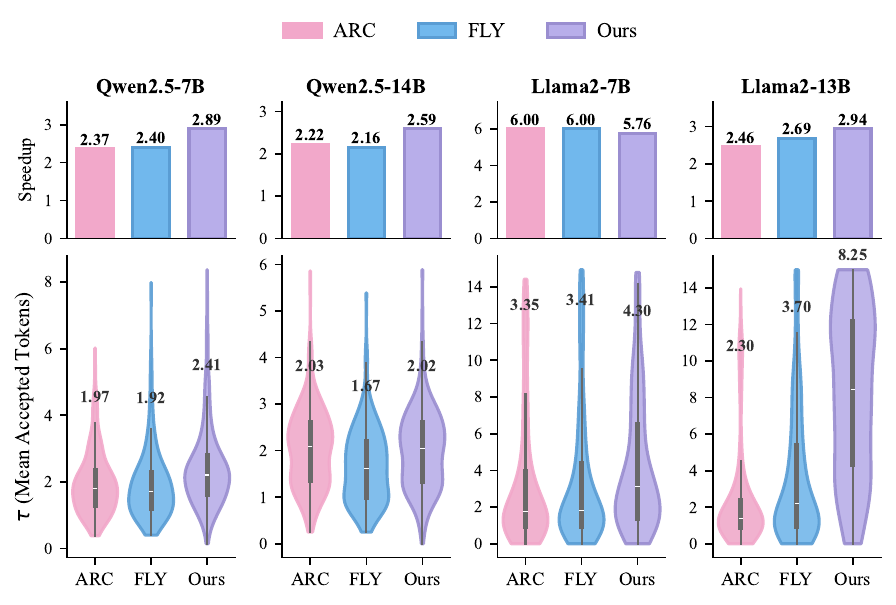}
    \caption{Verification method comparison. Upper: speedup; Lower: $\tau$. Our method outperforms baselines on most models.}
    \label{fig:verifier_comp}
\end{figure}

\subsection{Performance Comparison}
\textbf{Benchmarking against Training-Free Baselines.} \cref{tab:full_performance} shows \textit{SENSE(ID)} typically dominates by minimizing distributional divergence, with mean speedups of $\textbf{2.39}\text{--}\textbf{3.03}\times$. \textit{SENSE(OOD)} maintains high stability over other RSD methods. However, Llama2-7B presents a boundary condition where OOD prevails. We attribute this to the entropy-alignment trade-off: unlike confident strong models, weak models exhibit high intrinsic uncertainty, allowing OOD datastores to function as semantic correctives. Crucially, our entropy-based gating ($B_{ent}$) exploits this uncertainty to accept these corrective drafts, transforming generation noise into acceleration opportunities (see Appendix \cref{append:ID/OOD Boundary Conditions} for detailed boundary analysis).

\textbf{Comparison with Training-Based SOTA.} As detailed in \cref{tab:training_based_comp}, \textit{SENSE} effectively challenges the dominance of training-based paradigms. Despite a datastore construction time significantly lower than EAGLE's training overhead, \textit{SENSE} surpasses the SOTA method EAGL-2(3) across most metrics. Specifically, on Qwen3-14B,  \textit{SENSE} achieves superior mean speedups of $\textbf{2.98}\times$. Crucially, the consistently higher mean accepted tokens ($\tau$) validate that our retrieval-verification synergy generates higher-quality drafts than trained speculative heads, effectively bridging the efficiency gap between training-free and training-based approaches.

\textbf{Analysis of Verification Strategies.}
To isolate verification effects, we benchmark our Soft-gated Evaluation (SE) against SOTA relaxed verification baselines, ARC~\citep{noauthor_arc-decode_2025} and FLY~\citep{li_training-free_2025}, while keeping the retrieval module (\textit{SEN}) fixed. As shown in \cref{fig:verifier_comp}, \textit{SE} yields $\textbf{16--22}\%$ speedup gains on Qwen and Llama2-13B by salvaging semantic equivalents via synergistic masking ($B_{ent} \wedge B_{topk}$) that risk-bounded heuristics prune. The marginal exception on Llama2-7B aligns with the calibration constraints of weaker models discussed above, reinforcing that entropy-guided verification requires reliable probability estimates (comprehensive analysis in \cref{append:verifier_full}).

\begin{figure}[t]
    \centering
    \includegraphics[width=1\linewidth]{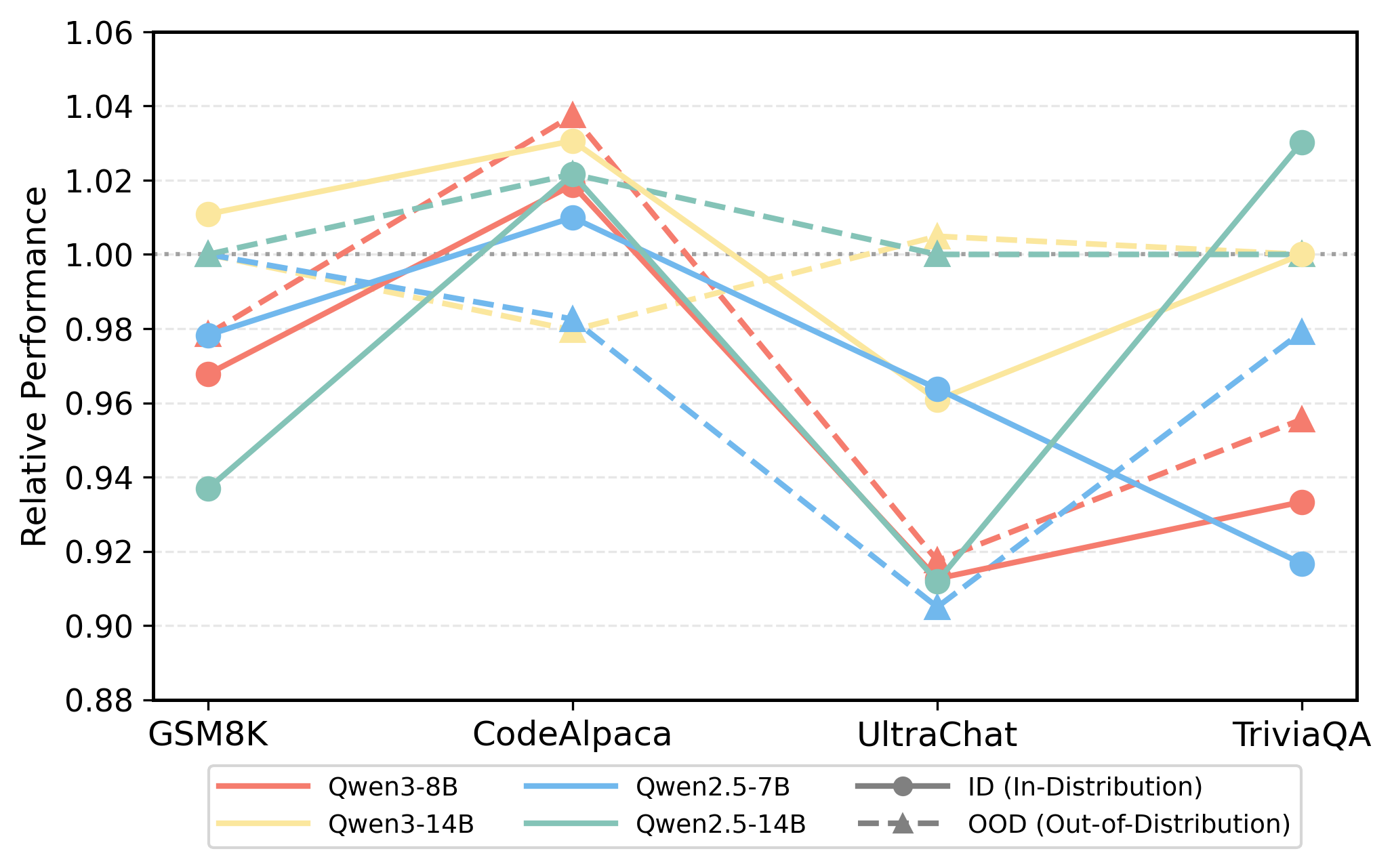}
    \caption{Generation quality preservation. \textit{SENSE} achieves comparable performance to the Vanilla across diverse domains.}
    \label{fig:accuracy}
\vspace{-1.5em}
\end{figure}


\textbf{Generation Quality Analysis.} Since \textit{SE} employs a losse verification mechanism, assessing the preservation of downstream task accuracy is critical. Consequently, we explicitly report accuracy preservation in \cref{fig:accuracy}. Quantitatively, our method maintains an average accuracy of \textbf{98.02\%} relative to the original target model across diverse tasks and varying model scales, demonstrating that our strategy ensures high output fidelity without significant degradation. Comprehensive numerical results are provided in \cref{append:accuracy}.

\begin{figure}[b]
\vspace{-1.5em}
    \centering
    \includegraphics[width=\columnwidth]{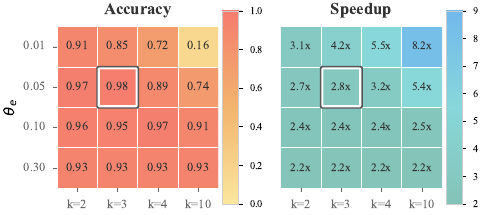}
    \caption{Hyperparameter sensitivity analysis on Qwen2.5-7B (GSM8K). Left: Accuracy (higher is better); Right: Speedup (higher is faster). Each cell shows the mean performance across window sizes $w \in \{5,6,7,8,9\}$. The highlighted cell indicates our default configuration ($\theta_e{=}0.05$, $k{=}3$), which achieves 98\% accuracy at 2.8$\times$ speedup.}
    \label{fig:hp_sensitivity}
\end{figure}

\subsection{Details Analysis}
\label{sec:3_ablation}

\begin{figure*}[t]
    \centering
    \includegraphics[width=1\linewidth]{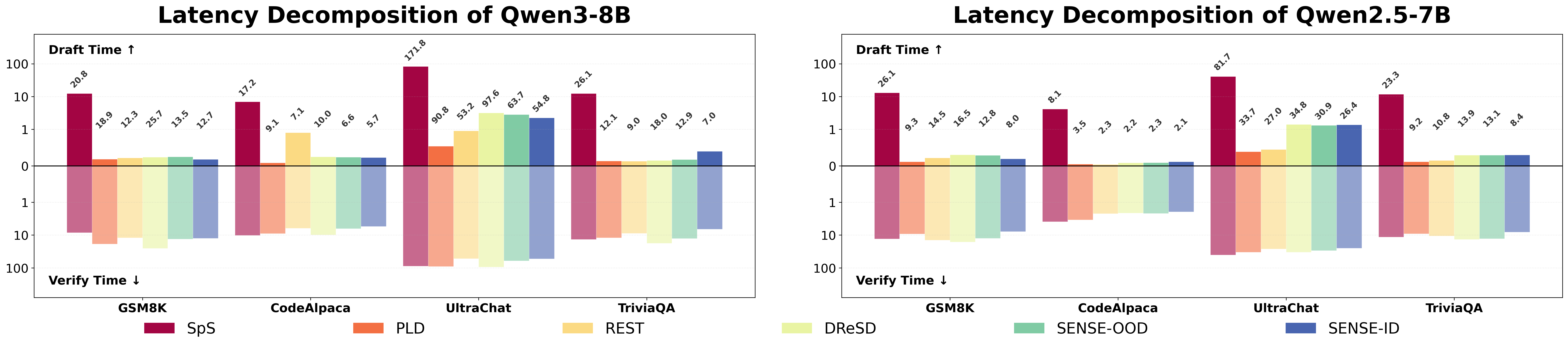}
    \caption{Latency Decomposition. Average latency per decoding step, decomposed into draft generation (upper) and verification (lower). Top annotations indicate total latency. \textbf{A symmetric log scale is used to accommodate the significant magnitude disparity.}}
    \label{fig:time_diff}
\vspace{-1em}
\end{figure*}

\begin{table}[b]
\centering
\caption{Component-wise ablation study of \textit{SENSE} using Qwen3-8B across four datasets. We evaluate the impact of the SEN and SE modules on Speedup and Mean Accepted Tokens ($\tau$).}
\label{tab:ablation}

\resizebox{\columnwidth}{!}{
\setlength{\tabcolsep}{2pt} 
\begin{tabular}{ll cc cc cc cc}
\toprule
\multirow{2.5}{*}{Datastore} & \multirow{2.5}{*}{Method} & \multicolumn{2}{c}{GSM8K}  & \multicolumn{2}{c}{CodeAlpaca} & \multicolumn{2}{c}{UltraChat} & \multicolumn{2}{c}{TriviaQA} \\

\cmidrule(lr){3-4} \cmidrule(lr){5-6} \cmidrule(lr){7-8} \cmidrule(lr){9-10}
& & Speedup & $\tau$ & Speedup & $\tau$ & Speedup & $\tau$ & Speedup & $\tau$ \\
\midrule

 \multirow{3}{*}{OOD} 
 & \textit{SENSE}         & \textbf{1.58} & \textbf{1.70} & \textbf{2.44} & \textbf{3.13} & \textbf{1.87} & \textbf{2.19} & \textbf{1.39} & \textbf{1.36} \\
 & \hspace{0.5em} w/o SEN & 1.14 & 1.35 & 2.16 & 2.34 & 1.36 & 1.41 & 1.24 & 1.24 \\
 & \hspace{0.5em} w/o SE  & 1.05 & 1.31 & 2.28 & 2.41 & 1.56 & 1.64 & 1.12 & 1.35 \\

\midrule
\multirow{3}{*}{ID} 
 & \textit{SENSE}         & \textbf{2.48} & \textbf{2.64} & \textbf{2.44} & \textbf{3.13} & \textbf{2.07} & \textbf{1.18} & \textbf{2.22} & \textbf{2.12} \\
 & \hspace{0.5em} w/o SEN & 2.07 & 2.20 & 2.16 & 2.34 & 1.86 & 1.70 & 1.97 & 1.84 \\
 & \hspace{0.5em} w/o SE  & 2.21 & 2.49 & 2.28 & 2.41 & 1.84 & 1.91 & 2.07 & 1.98 \\
\bottomrule
\end{tabular}
}
\end{table}
\textbf{Hyperparameter Sensitivity.} As shown in \cref{fig:hp_sensitivity}, we conduct a grid search on $\theta_e$, Top-$k$, and window size $w$, revealing a clear control hierarchy:
\textbf{(1) Primary Gate ($\theta_e$):} High values (0.30) block relaxed verification, collapsing all configurations to the baseline (Acc$=$0.93, Spd$=$2.2$\times$).
\textbf{(2) Trade-off Regulator ($k$):} At moderate $\theta_e$ (0.01--0.10), Top-$k$ dictates the balance. Large $k$ ($10$) yields extreme speed (8.2$\times$) but poor accuracy ($16\%$), while small $k$ ensures semantic alignment.
\textbf{(3) Negligible Factor ($w$):} Window size $w$ has minimal impact and is aggregated in the heatmap. Our default configuration ($\theta_e{=}0.05$, $k{=}3$) achieves the optimal accuracy-speedup balance. Detailed results are in \cref{tab:hp_full}.

\textbf{Ablation Study.} We conduct a component-wise analysis to isolate the distinct contribution of each module within \textit{SENSE}. As shown in \cref{tab:ablation}, ablating \textit{SEN}(\textit{SE}) leads to average speedup reductions of $\textbf{0.29}(\textbf{0.20})$ on ID and $\textbf{0.35}(\textbf{0.32})$ on OOD respectively. Specifically on GSM8K, ablating \textit{SEN} (\textit{SE}) causes speedups to drop to $\textbf{2.07}\times$ ($\textbf{2.21}\times$) on ID and $\textbf{1.14}\times$ ($\textbf{1.04}\times$) on OOD. This significant decay validates our design, where \textit{SEN} and \textit{SE} are mutually reinforcing under the framework of semantic alignment, balancing candidate diversity with verification flexibility.

\begin{figure}[t]
    \centering
    \includegraphics[width=1\linewidth]{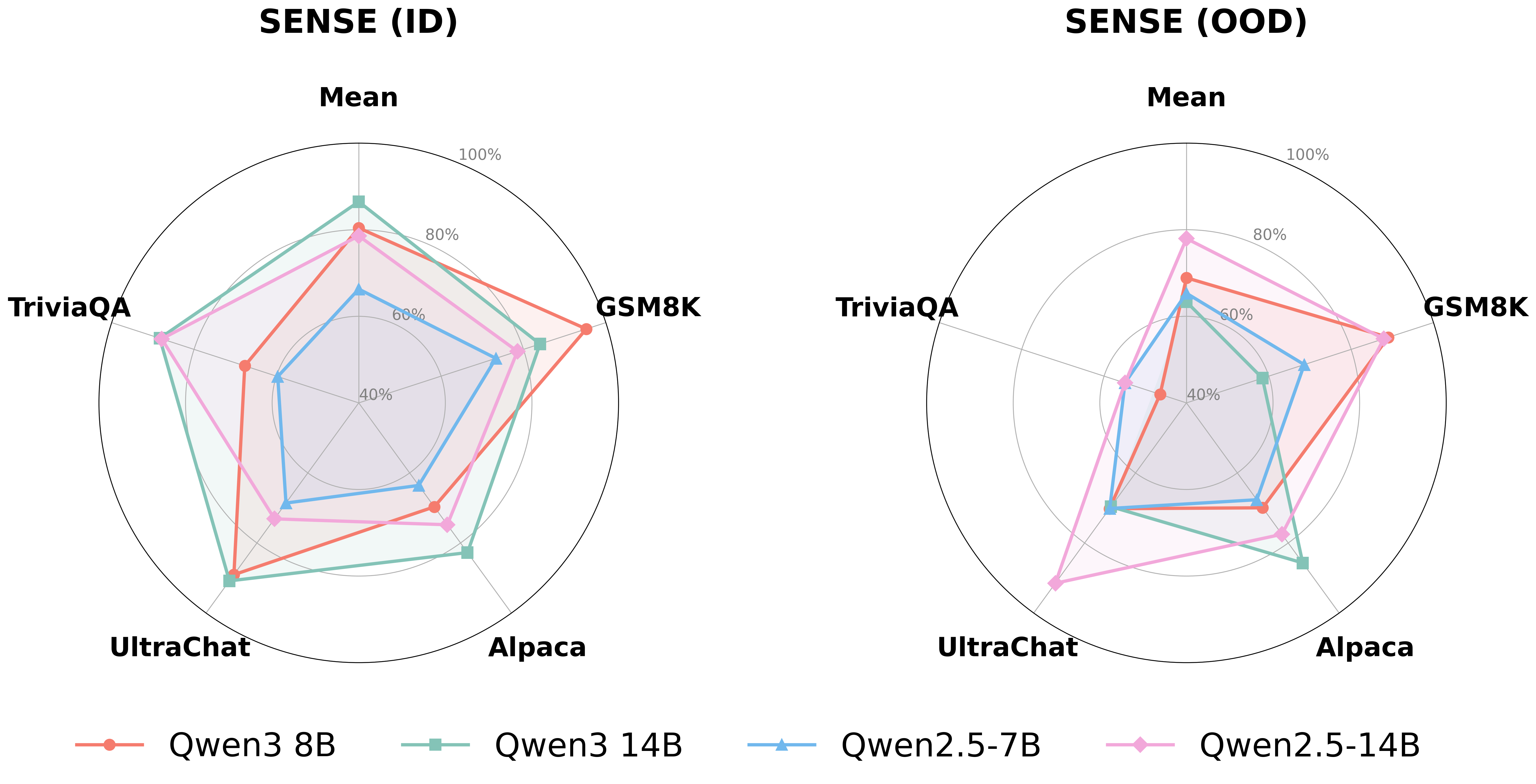}
    \caption{Compression efficiency analysis of the Losse Trie structure across diverse model scales and benchmarks.}
    \label{fig:trie_comp}
\vspace{-1.5em}
\end{figure}

\textbf{Losse Trie Compression Analysis.} As illustrated in \cref{fig:trie_comp}, our Loose Trie structure maintains robust storage efficiency, achieving an average compression ratio of $\textbf{73.4}\%$. Notably, compression efficiency exhibits an inverse correlation with model scale. This trend stems from the richer lexical diversity of larger models, where enhanced generative capabilities naturally diminish the local token redundancy required for effective compaction.


\textbf{Latency Decomposition Analysis.} As visualized in \cref{fig:time_diff}, we decompose the total latency into draft generation and verification phases.
By leveraging retrieval, \textit{SENSE} effectively bypasses the high drafting costs that burden generative baselines, reducing this overhead to negligible levels.
Moreover, within the verification phase, \textit{SENSE} exhibits superior efficiency and stability. This combination of minimal drafting latency and streamlined verification directly translates into superior end-to-end acceleration.


\section{Conclusion}
We present \textbf{\textit{SENSE}}, a novel retrieval-based speculative decoding framework anchored in semantic alignment to overcome lexical rigidity. 
To realize this philosophy, \textit{SENSE} leverages Semantic Embedding Navigation (\textit{SEN}) to augment retrieval diversity, and Soft-gated Evaluation (\textit{SE}) to enable tolerance-aware verification while preserving the target model's generation quality. 
To facilitate rigorous comparative analysis, we establish a unified evaluation platform that decomposes SD variants into atomic primitives. 
Extensive empirical evaluations across diverse benchmarks and backbones substantiate the effectiveness and robustness of \textit{SENSE}.
Furthermore, our results indicate that as model capabilities advance, the challenge of generating exact-match drafts will intensify, thereby underscoring the pivotal role of semantic-based verification in future developments.




\section*{Impact Statement}

This paper presents work whose goal is to advance the field of Machine
Learning. There are many potential societal consequences of our work, none
which we feel must be specifically highlighted here.



\bibliography{example_paper}
\bibliographystyle{icml2026}

\newpage
\appendix
\onecolumn
\section{Outline}

This appendix provides supplementary materials to complement the main text:

\begin{itemize}
    \item \hyperref[sec:appendix_llm]{Section~\ref*{sec:appendix_llm}} describes the usage of large language models during manuscript preparation.
    
    \item \hyperref[sec:appendix_algorithm]{Section~\ref*{sec:appendix_algorithm}} presents the detailed algorithmic specifications of the \textit{SENSE} framework.
    
    \item \hyperref[sec:appendix_analysis]{Section~\ref*{sec:appendix_analysis}} provides extended experimental analyses, including  ID/OOD boundary conditions analysis, complete verification strategy comparisons, comprehensive hyperparameter study, accuracy preservation analysis, a case study demonstrating output fidelity and speedup, a failure case analysis,as well as detailed datastore resources statics.
    
    \item \hyperref[sec:appendix_details]{Section~\ref*{sec:appendix_details}} offers comprehensive experimental details, covering dataset statistics, prompt templates, baseline descriptions, implementation hyperparameters, and hardware configurations.
    
    \item \hyperref[sec:appendix_future]{Section~\ref*{sec:appendix_future}} outlines promising directions for future research.
\end{itemize}

\section{LLM Usage}
\label{sec:appendix_llm}

The large language model (LLM) was utilized as a writing assistant during the preparation of this manuscript. Its application was strictly limited to improving the clarity and grammatical accuracy of the text. Specific uses included rephrasing sentences for better flow and translating initial concepts and drafts from Chinese to English. All core scientific contributions, including the conceptualization of our \textit{\textbf{SENSE}} framework, the design of the methodology and experiments, and the analysis and interpretation of the results, are solely the work of the authors. The authors take full responsibility for all claims and the final content of this paper.

\section{Algorithm Details}
\label{sec:appendix_algorithm}

This section provides the complete algorithmic specification of the \textit{SENSE} framework. We present four algorithms: the main speculative decoding loop (Algorithm~\ref{alg:sense_main}), the Semantic Embedding Navigation module (Algorithm~\ref{alg:sen}), the Loose Trie construction (Algorithm~\ref{alg:loose_trie}), and the Soft-gated Evaluation mechanism (Algorithm~\ref{alg:sge}).


\subsection{Main Decoding Loop}

Algorithm~\ref{alg:sense_main} presents the main speculative decoding loop. The framework operates in three phases per iteration: (1) draft generation via semantic retrieval, (2) parallel verification with tree-structured attention, and (3) token acceptance with state updates.

\begin{algorithm}[t]
\caption{SENSE Framework}
\label{alg:sense_main}
\begin{algorithmic}[1]

\REQUIRE Prompt $q$, target model $\mathcal{M}$, retriever $\mathcal{R}$, verifier $\mathcal{V}$, max length $T$
\ENSURE Generated sequence $\mathbf{y}$

\STATE

\STATE $\mathbf{y} \leftarrow \textsc{Tokenize}(q)$
\STATE $n \leftarrow |\mathbf{y}|$

\STATE

\STATE $h, p, \mathcal{K} \leftarrow \mathcal{M}(\mathbf{y}_{1:n})$
\STATE $\mathcal{R}.\textsc{Update}(h, \mathbf{y})$

\STATE

\WHILE{$|\mathbf{y}| < T$ \textbf{and} $\mathbf{y}_{-1} \neq \texttt{EOS}$}

    \STATE

    \STATE $\tilde{x} \leftarrow \arg\max p$
    \STATE $D, \mathcal{T}, \mathbf{M}, \Gamma \leftarrow \textsc{SEN}(h, \tilde{x}, \mathcal{R})$

    \STATE

    \STATE $\mathbf{M}_{\text{full}} \leftarrow \textsc{TreeMask}(\mathbf{M}, |\mathbf{y}|, |\mathcal{K}|)$
    \STATE $H', P' \leftarrow \mathcal{M}.\textsc{Forward}(\mathcal{T}, \mathbf{M}_{\text{full}}, \mathcal{K})$
    \STATE $H_{\text{align}}, P_{\text{align}} \leftarrow [h; \Gamma(H')], [p; \Gamma(P')]$
    \STATE $i^*, L^* \leftarrow \textsc{SGE}(D, P_{\text{align}}, \mathcal{V})$

    \STATE

    \IF{$L^* = 0$}
        \STATE $\mathbf{a} \leftarrow [D_{i^*, 0}]$
    \ELSE
        \STATE $\mathbf{a} \leftarrow D_{i^*, 0:L^*+1}$
    \ENDIF

    \STATE

    \STATE $\mathbf{y} \leftarrow \mathbf{y} \| \mathbf{a}$
    \STATE $h \leftarrow H_{\text{align}}[i^*, L^*+1]$
    \STATE $p \leftarrow P_{\text{align}}[i^*, L^*+1]$
    \STATE $\mathcal{K} \leftarrow \textsc{UpdateKV}(\mathcal{K}, i^*, \Gamma)$
    \STATE $\mathcal{R}.\textsc{Update}(H_{\text{align}}[i^*, 0:L^*+1], \mathbf{a})$

\ENDWHILE

\STATE

\STATE \textbf{Return} $\mathbf{y}$

\end{algorithmic}
\end{algorithm}

The prefill phase (lines 2--6) initializes the generation by encoding the prompt and populating the dynamic retrieval cache. Each decoding iteration consists of: draft generation (lines 9--10) using Algorithm~\ref{alg:sen}, tree-structured verification (lines 12--15) with a single forward pass over the flattened candidates, and state update (lines 17--24) where accepted tokens are appended to the output sequence.


\subsection{Semantic Embedding Navigation}

Algorithm~\ref{alg:sen} describes the draft generation module. It retrieves candidate continuations from both a static corpus and a dynamic in-context cache, then organizes them into a prefix-compressed tree structure for efficient parallel verification.

\begin{algorithm}[t]
\caption{Semantic Embedding Navigation (SEN)}
\label{alg:sen}
\begin{algorithmic}[1]

\REQUIRE Hidden state $h \in \mathbb{R}^d$, anchor token $\tilde{x}$, retriever $\mathcal{R}$
\REQUIRE Hyperparameters: $N$, $M$, $N'$, $\alpha$, $\beta$ where $\alpha \gg \beta$
\ENSURE Candidates $D$, flattened tree $\mathcal{T}$, mask $\mathbf{M}$, mapping $\Gamma$

\STATE

\STATE $k \leftarrow \textsc{L2Norm}(\textsc{PCA}(h))$

\STATE

\STATE $D_s, \mathbf{s}_s \leftarrow \textsc{FAISS}(k, \mathcal{C}_s, N')$
\STATE $D_d, \mathbf{s}_d \leftarrow \textsc{DynSearch}(h, \mathcal{C}_d, \tilde{x})$
\STATE $D_{\text{all}} \leftarrow D_s \cup D_d$
\STATE $\mathbf{s} \leftarrow [\mathbf{s}_s; \mathbf{s}_d + \delta]$

\STATE

\FOR{$i = 1$ \textbf{to} $|D_{\text{all}}|$}
    \STATE $S_i \leftarrow \alpha \cdot \mathbb{I}(D_{\text{all}}[i, 0] = \tilde{x}) + \beta \cdot \mathbf{s}_i$
\ENDFOR
\STATE $D \leftarrow \textsc{TopN}(D_{\text{all}}, S)$
\STATE $D[0, 0] \leftarrow \tilde{x}$
\STATE $D[-1, 0] \leftarrow \tilde{x}$

\STATE

\STATE $D_{\text{sorted}} \leftarrow \textsc{LexSort}(D)$
\STATE $\mathcal{T}, \mathbf{M}, \Gamma \leftarrow \textsc{LooseTrie}(D_{\text{sorted}})$

\STATE

\STATE \textbf{Return} $D, \mathcal{T}, \mathbf{M}, \Gamma$

\end{algorithmic}
\end{algorithm}

The module operates in four steps:

\paragraph{Query Construction (line 2).} The hidden state $h$ is projected to a lower-dimensional space via PCA and L2-normalized to form the retrieval query $k \in \mathbb{R}^v$ where $v \ll d$.

\paragraph{Hybrid Retrieval (lines 4--7).} Candidates are retrieved from two sources: a static pre-built corpus $\mathcal{C}_s$ via FAISS approximate nearest neighbor search, and a dynamic cache $\mathcal{C}_d$ that stores recent generation context. Dynamic results receive a score boost $\delta$ to prioritize in-context continuations.

\paragraph{Composite Scoring (lines 9--14).} Each candidate is scored by a weighted combination of lexical matching (whether its first token equals the anchor $\tilde{x}$) and semantic similarity. The constraint $\alpha \gg \beta$ ensures candidates with matching first tokens are strongly preferred. Lines 13--14 guarantee at least two candidates start with the anchor token.

\paragraph{Trie Construction (lines 16--17).} Selected candidates are lexicographically sorted and passed to Algorithm~\ref{alg:loose_trie} for prefix compression.


\subsection{Loose Trie Construction}

Algorithm~\ref{alg:loose_trie} constructs a prefix-compressed tree from sorted candidates. Unlike a canonical trie that requires $O(NM^2)$ pairwise comparisons, our approach exploits lexicographic ordering to achieve $O(NM)$ complexity through adjacent-only comparisons.

\begin{algorithm}[t]
\caption{Loose Trie Construction}
\label{alg:loose_trie}
\begin{algorithmic}[1]

\REQUIRE Sorted candidates $D \in \mathbb{Z}^{N \times M}$
\ENSURE Flattened tree $\mathcal{T}$, attention mask $\mathbf{M}$, mapping $\Gamma$

\STATE

\STATE $\texttt{match} \leftarrow \textsc{Zeros}(N, M)$
\STATE $\texttt{match}[1:] \leftarrow (D[1:] = D[:-1])$
\STATE $\texttt{is\_prefix} \leftarrow \textsc{CumProd}(\texttt{match}, \text{dim}=1)$
\STATE $\texttt{unique} \leftarrow \neg\, \texttt{is\_prefix}$

\STATE

\STATE $\mathcal{T} \leftarrow D[\texttt{unique}]$
\STATE $L \leftarrow |\mathcal{T}|$

\STATE

\STATE $\texttt{idx} \leftarrow \textsc{CumSum}(\texttt{unique}.\textsc{Flat}()) - 1$
\STATE $\Gamma \leftarrow \textsc{Reshape}(\texttt{idx}, (N, M))$
\STATE $\texttt{tmp} \leftarrow \textsc{Where}(\texttt{unique}, \Gamma, -1)$
\STATE $\Gamma \leftarrow \textsc{CumMax}(\texttt{tmp}, \text{dim}=0)$

\STATE

\STATE $\mathbf{M} \leftarrow \textsc{Zeros}(L, L)$
\STATE $\mathbf{I}, \mathbf{J} \leftarrow \textsc{TrilIndices}(M, M)$
\FOR{$i = 1$ \textbf{to} $N$}
    \STATE $\mathbf{M}[\Gamma[i, \mathbf{I}], \Gamma[i, \mathbf{J}]] \leftarrow 1$
\ENDFOR

\STATE

\STATE \textbf{Return} $\mathcal{T}, \mathbf{M}, \Gamma$

\end{algorithmic}
\end{algorithm}

The algorithm proceeds as follows:

\paragraph{Prefix Detection (lines 2--5).} For each position, we check whether the current candidate matches its predecessor (line 3). The cumulative product along dimension 1 (line 4) propagates this match status: a position is marked as shared prefix only if all preceding positions also match. The negation yields the unique token mask.

\paragraph{Tree Extraction (lines 7--8).} Tokens marked as unique are extracted to form the flattened tree $\mathcal{T}$, eliminating redundant prefix tokens.

\paragraph{Index Mapping (lines 10--13).} The gather-scatter mapping $\Gamma$ records which tree position each candidate token maps to. For shared prefixes, indices are propagated from predecessor candidates via cumulative maximum.

\paragraph{Attention Mask (lines 15--19).} The tree attention mask $\mathbf{M}$ encodes causal visibility: within each candidate path, earlier positions can be attended to by later positions, following standard causal attention semantics.


\subsection{Soft-gated Evaluation}

Algorithm~\ref{alg:sge} performs verification through a cascade of four binary masks. This design allows relaxed acceptance of semantically plausible tokens when the model exhibits high uncertainty.

\begin{algorithm}[t]
\caption{Soft-gated Evaluation (SGE)}
\label{alg:sge}
\begin{algorithmic}[1]

\REQUIRE Candidates $D \in \mathbb{Z}^{N \times M}$, logits $P \in \mathbb{R}^{N \times M \times |\mathcal{V}|}$
\REQUIRE Hyperparameters: threshold $\gamma$, top-$k$, window $W$
\ENSURE Best path $i^*$, acceptance length $L^*$

\STATE

\STATE $Y \leftarrow \arg\max_v P[:, :, v]$

\STATE

\STATE $B_{\text{em}} \leftarrow \mathbb{I}(D = Y)$

\STATE

\STATE $\tilde{P} \leftarrow \textsc{Softmax}(P)$
\STATE $E \leftarrow -\sum_v \tilde{P}_{:,:,v} \log \tilde{P}_{:,:,v}$
\STATE $B_{\text{ent}} \leftarrow \mathbb{I}(E / \log|\mathcal{V}| > \gamma)$

\STATE

\STATE $\mathcal{Y}_k \leftarrow \textsc{TopK}(P, k)$
\STATE $B_{\text{topk}} \leftarrow \mathbb{I}(D \in \mathcal{Y}_k)$

\STATE

\STATE $\bar{B} \leftarrow \textsc{Pad}(1 - B_{\text{em}}, (0, W), 1)$
\STATE $R \leftarrow \textsc{Conv1D}(\bar{B}, \mathbf{1}_W)[:, 1:M+1]$
\STATE $B_{\text{con}} \leftarrow \mathbb{I}(R = 0)$

\STATE

\STATE $G \leftarrow B_{\text{em}} \lor (B_{\text{ent}} \land (B_{\text{topk}} \lor B_{\text{con}}))$

\STATE

\STATE $G_{\text{seq}} \leftarrow \textsc{CumProd}(G, \text{dim}=1)$
\STATE $L_i \leftarrow \sum_j G_{\text{seq}}[i, j] - 1$
\STATE $i^* \leftarrow \arg\max_i L_i$
\STATE $L^* \leftarrow L_{i^*}$

\STATE

\STATE \textbf{Return} $i^*, L^*$

\end{algorithmic}
\end{algorithm}

The four masks capture different acceptance criteria:

\paragraph{Exact Match $B_{\text{em}}$ (line 4).} The baseline criterion: accept if the draft token exactly matches the greedy prediction.

\paragraph{Entropy Gate $B_{\text{ent}}$ (lines 6--8).} Identifies high-uncertainty positions where the model's probability mass is spread across multiple tokens. Normalized entropy exceeding threshold $\gamma$ indicates the model considers multiple continuations plausible.

\paragraph{Top-$k$ Relaxation $B_{\text{topk}}$ (lines 10--11).} For uncertain positions, accept tokens within the top-$k$ predictions. This captures semantically equivalent alternatives that the model assigns high probability.

\paragraph{Convolutional Lookahead $B_{\text{con}}$ (lines 13--15).} A robustness check examining whether mismatches are isolated or clustered. The 1D convolution counts mismatches in a sliding window of size $W$. Isolated mismatches (zero future mismatches) are more likely benign.

\paragraph{Final Decision (line 17).} A token is accepted if it matches exactly, \emph{or} if the position has high uncertainty \emph{and} the token satisfies either the top-$k$ or lookahead criterion. The cumulative product (line 19) ensures only contiguous prefixes are accepted.


\subsection{Complexity Analysis}

Let $N$ denote the number of draft candidates, $M$ the maximum draft length, and $|\mathcal{C}_s|$ the static corpus size.

\begin{itemize}
    \item \textbf{Retrieval:} $O(\log|\mathcal{C}_s|)$ for FAISS search, $O(|\mathcal{C}_d|)$ for dynamic cache lookup.
    \item \textbf{Trie Construction:} $O(NM)$ via vectorized adjacent comparisons.
    \item \textbf{Verification:} $O(NM)$ through fully parallelized mask operations.
    \item \textbf{Model Forward:} Single pass over $L \leq NM$ flattened tree tokens.
\end{itemize}

The dominant cost per iteration is the target model forward pass, whose complexity is reduced from $O(NM)$ separate calls to a single batched call through tree-structured attention.


\subsection{Tree Attention Mask for Static KV Cache}

Given the draft attention mask $\mathbf{M} \in \{0, 1\}^{L \times L}$, current sequence length $m$, and maximum cache length $M_{\text{cache}}$, we construct:
\begin{equation}
\mathbf{M}_{\text{full}}[i, j] =
\begin{cases}
-\infty & \text{otherwise} \\
0 & \text{if } j < m \\
0 & \text{if } m \leq j < m+L \text{ and }\mathbf{M}[i, j-m] = 1\\
\end{cases}
\end{equation}

\section{Further Analysis}
\label{sec:appendix_analysis}

\subsection{Analysis on ID/OOD Boundary Conditions}
\label{append:ID/OOD Boundary Conditions}
The inversion observed in Llama2-7B (where OOD outperforms ID) highlights a critical boundary condition driven by model confidence.

\begin{itemize}
    \item Low-Entropy Regime (Strong Models): Models like Qwen3 are highly confident (low entropy). Any deviation in the OOD datastore is treated as a distribution shift and rejected. Thus, ID (distributional alignment) is essential.
\end{itemize}

\begin{itemize}
    \item High-Entropy Regime (Weak Models): Models like Llama2-7B exhibit high uncertainty and generation noise. In this regime, ID datastores merely mimic the model's own hallucinations. OOD datastores, derived from ground truth, provide better semantic guidance. SENSE's entropy-gating mechanism is uniquely designed to bypass exact matching in these high-uncertainty regions, allowing the model to accept these "corrective" OOD drafts.
\end{itemize}

\subsection{Verification Strategy Comparison}
\label{append:verifier_full}

To provide a comprehensive view of verification strategy performance, \cref{tab:verifier_comp_full} presents the complete comparison across all model-dataset combinations. We fix the SEN retrieval module and compare different verification methods: ARC, FLY, and our Soft-gated Evaluation (SE). The table reports both speedup ratios and mean accepted tokens ($\tau$) for each configuration.

The superiority of SE stems from its synergistic masking mechanism $G = B_{em} \vee (B_{ent} \wedge (B_{topk} \vee B_{con}))$.

Unlike ARC, which rejects candidates solely based on a rigid risk bound, or FLY, which relies on a strictly linear lookahead, SE introduces a \textit{dual-pathway rescue mechanism}:

\begin{enumerate}
    \item \textbf{Distributional Rescue ($B_{topk}$):} For high-entropy positions where the model is uncertain (e.g., synonyms like "glad" vs. "happy"), SE accepts the token if it resides within the top-$k$ probability mass. This captures semantic equivalence that exact-match or strict risk bounds often miss.
    
    \item \textbf{Structural Rescue ($B_{con}$):} For positions with isolated lexical mismatches (e.g., singular/plural variations), the convolutional mask ($B_{con}$) aggregates local error density. If the mismatch is an isolated event within a correct neighborhood, the token is retained.
\end{enumerate}

This combination allows SE to salvage valid drafts that baselines would erroneously prune, directly translating to higher acceptance rates and speedups.

\textbf{The Calibration Boundary Condition (Llama2-7B).} 

As noted in \cref{append:verifier_full}, baselines exhibit a marginal lead on Llama2-7B (e.g., 6.00$\times$ vs. 5.85$\times$). While \cref{append:ID/OOD Boundary Conditions} attributes the success of OOD datastores to the \textit{generation quality} of weak models, the verification performance is governed by \textit{model calibration}.

Entropy-guided verification operates on the assumption that \textit{predicted uncertainty reflects semantic flexibility}. However, smaller, older models like Llama2-7B often suffer from poor calibration—they may exhibit "confident errors" (low entropy on wrong tokens) or "diffuse correctness" (high entropy on the correct token but poor top-$k$ clustering).

In such regimes, the gating signal $B_{ent}$ becomes noisy, occasionally triggering relaxed verification when it should not, or failing to trigger it when appropriate. In contrast, modern architectures (Qwen series) demonstrate superior calibration, making entropy a reliable proxy for semantic verification and enabling SE to reach its full potential.

\begin{table*}[t]
\renewcommand{\arraystretch}{0.85}
\centering
\small
\caption{Complete performance comparison of verification strategies. We fix the SEN retrieval module and compare different verification methods: ARC, FLY, and our Soft-gated Evaluation (SE). \textbf{Bold} indicates the best performance in each row.}
\label{tab:verifier_comp_full}

\setlength{\tabcolsep}{3.5pt}
\begin{tabular}{ll cc cc cc cc cc}
\toprule
\multirow{2.5}{*}{Model} & \multirow{2.5}{*}{Method} & \multicolumn{2}{c}{GSM8K} & \multicolumn{2}{c}{CodeAlpaca} & \multicolumn{2}{c}{UltraChat} & \multicolumn{2}{c}{TriviaQA} & \multicolumn{2}{c}{Mean} \\
\cmidrule(lr){3-4} \cmidrule(lr){5-6} \cmidrule(lr){7-8} \cmidrule(lr){9-10} \cmidrule(lr){11-12}
& & Speedup & $\tau$ & Speedup & $\tau$ & Speedup & $\tau$ & Speedup & $\tau$ & Speedup & $\tau$ \\
\midrule
\multirow{3}{*}{Qwen2.5-7B}
 & ARC & 2.51 & 3.09 & 2.87 & 3.71 & 2.14 & 2.71 & 1.97 & 2.42 & 2.37 & 2.98 \\
 & FLY & 2.43 & 3.02 & 2.83 & 3.71 & 2.02 & 2.57 & 2.32 & 2.89 & 2.40 & 3.05 \\
 & \textbf{Ours} & \textbf{2.94} & \textbf{3.44} & \textbf{3.30} & \textbf{3.95} & \textbf{2.59} & \textbf{3.04} & \textbf{2.72} & \textbf{3.63} & \textbf{2.89} & \textbf{3.52} \\
\midrule
\multirow{3}{*}{Qwen2.5-14B}
 & ARC & 2.66 & 3.26 & \textbf{2.49} & 3.12 & 1.89 & 1.33 & 1.85 & 2.28 & 2.22 & 2.50 \\
 & FLY & 2.51 & 3.10 & \textbf{2.49} & \textbf{3.14} & 1.77 & 2.20 & 1.87 & 2.32 & 2.16 & 2.69 \\
 & \textbf{Ours} & \textbf{3.00} & \textbf{3.48} & 2.48 & 2.90 & \textbf{2.40} & \textbf{2.98} & \textbf{2.47} & \textbf{2.56} & \textbf{2.59} & \textbf{2.96} \\
\midrule
\multirow{3}{*}{Llama2-7B}
 & ARC & 6.73 & 7.00 & 6.34 & 5.17 & \textbf{8.52} & \textbf{9.08} & 2.41 & 2.20 & 6.00 & 5.86 \\
 & FLY & \textbf{7.40} & \textbf{7.82} & \textbf{6.92} & \textbf{5.73} & 6.78 & 7.09 & 2.88 & 2.62 & \textbf{6.00} & 5.82 \\
 & \textbf{Ours} & 5.45 & 5.80 & 6.86 & 5.63 & 3.28 & 3.44 & \textbf{7.46} & \textbf{7.78} & 5.76 & \textbf{5.85} \\
\midrule
\multirow{3}{*}{Llama2-13B}
 & ARC & 1.14 & 2.29 & 2.60 & 4.79 & 1.22 & 2.31 & 4.89 & 8.94 & 2.46 & 4.58 \\
 & FLY & 1.10 & 2.26 & 3.36 & 6.19 & 1.49 & \textbf{2.83} & 4.81 & 8.75 & 2.69 & 5.01 \\
 & \textbf{Ours} & \textbf{1.24} & \textbf{2.44} & \textbf{3.72} & \textbf{6.83} & \textbf{1.45} & 2.73 & \textbf{5.36} & \textbf{9.75} & \textbf{2.94} & \textbf{5.44} \\
\bottomrule
\end{tabular}
\end{table*}

\subsection{Comprehensive Hyperparameter Study}
\label{append:hp_study}

To provide a thorough understanding of SENSE's sensitivity to hyperparameter choices, we conduct an exhaustive grid search over the three key parameters in the Soft-gated Evaluation module: entropy threshold ($\theta_e$), Top-$k$ value, and convolution window size ($w$). All experiments are conducted on Qwen2.5-7B using the GSM8K benchmark with the ID datastore.

\paragraph{Parameter Ranges.} We evaluate:
\begin{itemize}[noitemsep,topsep=0pt]
    \item Entropy threshold $\theta_e \in \{0.01, 0.05, 0.10, 0.30\}$
    \item Top-$k \in \{2, 3, 4, 10\}$
    \item Window size $w \in \{5, 6, 7, 8, 9\}$
\end{itemize}
This yields $4 \times 4 \times 5 = 80$ configurations. Complete results are presented in \cref{tab:hp_full}.

\paragraph{Key Observations.}
(1) \textbf{Entropy threshold} acts as the primary gating mechanism: higher values ($\theta_e = 0.30$) restrict relaxed verification to only the most uncertain positions, yielding conservative but stable performance; lower values ($\theta_e = 0.01$) enable more aggressive acceptance, improving speedup but requiring careful tuning of downstream criteria.
(2) \textbf{Top-$k$} controls the distributional tolerance: smaller $k$ values enforce stricter semantic alignment, while larger values ($k = 10$) risk accepting low-probability tokens that may degrade accuracy.
(3) \textbf{Window size} affects robustness to isolated mismatches: larger windows provide more context for the convolutional lookahead but may over-smooth the acceptance signal.

\begin{table*}[t]
\centering
\caption{Comprehensive hyperparameter study on Qwen2.5-7B (GSM8K, ID datastore). We report Accuracy (\%), Speedup ratio, and mean accepted tokens ($\tau$) across 80 configurations of entropy threshold $\theta_e$, Top-$k$, and window size $w$. \colorbox{gray!20}{Highlighted cells} indicate our default configuration ($\theta_e{=}0.05$, $k{=}3$, $w{=}6$).}
\label{tab:hp_full}
\small
\renewcommand{\arraystretch}{0.95}
\resizebox{\textwidth}{!}{
\setlength{\tabcolsep}{4pt}
\begin{tabular}{cc ccc ccc ccc ccc ccc}
\toprule
& & \multicolumn{3}{c}{$k=2$} & \multicolumn{3}{c}{$k=3$} & \multicolumn{3}{c}{$k=4$} & \multicolumn{3}{c}{$k=10$} \\
\cmidrule(lr){3-5} \cmidrule(lr){6-8} \cmidrule(lr){9-11} \cmidrule(lr){12-14}
$\theta_e$ & $w$ & Acc & SpeedUp & $\tau$ & Acc & SpeedUp & $\tau$ & Acc & SpeedUp & $\tau$ & Acc & SpeedUp & $\tau$ \\
\midrule
\multirow{5}{*}{0.01}
 & 5  & 0.89 & 3.28 & 3.93 & 0.76 & 5.02 & 5.70 & 0.67 & 5.95 & 6.50 & 0.15 & 8.12 & 8.28 \\
 & 6  & 0.90 & 3.25 & 3.85 & 0.82 & 4.30 & 4.99 & 0.77 & 5.59 & 6.14 & 0.12 & 8.33 & 8.46 \\
 & 7  & 0.90 & 3.07 & 3.61 & 0.89 & 3.89 & 4.53 & 0.73 & 5.09 & 5.64 & 0.16 & 8.16 & 8.35 \\
 & 8  & 0.92 & 3.00 & 3.55 & 0.88 & 3.93 & 4.57 & 0.71 & 5.23 & 5.79 & 0.18 & 8.12 & 8.31 \\
 & 9  & 0.93 & 2.95 & 3.51 & 0.88 & 3.96 & 4.59 & 0.72 & 5.60 & 6.15 & 0.18 & 8.10 & 8.27 \\
\midrule
\multirow{5}{*}{0.05}
 & 5  & 0.99 & 2.74 & 3.36 & 0.96 & 2.98 & 3.59 & 0.87 & 3.19 & 3.85 & 0.70 & 5.81 & 6.39 \\
 & 6  & 0.97 & 2.70 & 3.29 & \cellcolor{gray!20}0.98 & \cellcolor{gray!20}2.82 & \cellcolor{gray!20}3.44 & 0.88 & 3.42 & 4.10 & 0.73 & 5.62 & 6.19 \\
 & 7  & 0.97 & 2.67 & 3.27 & 0.97 & 2.79 & 3.40 & 0.89 & 3.14 & 3.77 & 0.74 & 5.28 & 5.91 \\
 & 8  & 0.97 & 2.67 & 3.25 & 0.97 & 2.79 & 3.39 & 0.89 & 3.16 & 3.75 & 0.76 & 5.28 & 5.93 \\
 & 9  & 0.97 & 2.65 & 3.23 & 0.98 & 2.75 & 3.37 & 0.90 & 3.17 & 3.75 & 0.78 & 5.09 & 5.73 \\
\midrule
\multirow{5}{*}{0.10}
 & 5  & 0.95 & 2.37 & 2.95 & 0.95 & 2.41 & 2.98 & 0.95 & 2.42 & 3.00 & 0.90 & 2.48 & 3.06 \\
 & 6  & 0.96 & 2.36 & 2.93 & 0.95 & 2.39 & 2.97 & 0.96 & 2.43 & 3.00 & 0.91 & 2.46 & 3.06 \\
 & 7  & 0.96 & 2.35 & 2.92 & 0.95 & 2.40 & 2.96 & 0.97 & 2.41 & 2.98 & 0.91 & 2.47 & 3.06 \\
 & 8  & 0.96 & 2.36 & 2.92 & 0.95 & 2.40 & 2.96 & 0.98 & 2.40 & 2.97 & 0.91 & 2.47 & 3.05 \\
 & 9  & 0.96 & 2.35 & 2.91 & 0.95 & 2.38 & 2.94 & 0.98 & 2.39 & 2.97 & 0.91 & 2.46 & 3.05 \\
\midrule
\multirow{5}{*}{0.30}
 & 5  & 0.93 & 2.24 & 2.78 & 0.93 & 2.25 & 2.78 & 0.93 & 2.24 & 2.78 & 0.93 & 2.24 & 2.78 \\
 & 6  & 0.93 & 2.25 & 2.78 & 0.93 & 2.25 & 2.78 & 0.93 & 2.25 & 2.78 & 0.93 & 2.24 & 2.78 \\
 & 7  & 0.93 & 2.24 & 2.78 & 0.93 & 2.23 & 2.78 & 0.93 & 2.26 & 2.78 & 0.93 & 2.26 & 2.78 \\
 & 8  & 0.93 & 2.24 & 2.78 & 0.93 & 2.26 & 2.78 & 0.93 & 2.24 & 2.78 & 0.93 & 2.25 & 2.78 \\
 & 9  & 0.93 & 2.25 & 2.78 & 0.93 & 2.26 & 2.78 & 0.93 & 2.25 & 2.78 & 0.93 & 2.24 & 2.78 \\
\bottomrule
\end{tabular}
}
\end{table*}

\subsection{Accuracy Preservation Analysis}
\label{append:accuracy}

To validate that SENSE preserves generation quality while achieving speedup, we conduct a comprehensive accuracy ablation study across multiple model-dataset configurations. Table~\ref{tab:accuracy_ablation} reports the task-specific metrics for vanilla (unaccelerated) decoding versus SENSE with both OOD and ID datastores. For GSM8K and TriviaQA, we report exact-match accuracy (ACC); for CodeAlpaca and UltraChat, we report ROUGE-L scores to assess generation quality.

\begin{table}[t]
\renewcommand{\arraystretch}{0.8}
\centering
\caption{Accuracy preservation analysis across models and datasets. We compare vanilla decoding (baseline) against SENSE-OOD and SENSE-ID configurations. GSM8K and TriviaQA report accuracy (ACC); CodeAlpaca and UltraChat report ROUGE-L scores. \textbf{Bold} indicates highest value per model-dataset pair.}
\label{tab:accuracy_ablation}

\setlength{\tabcolsep}{5pt}
\begin{tabular}{ll cccc}
\toprule
\multirow{2.5}{*}{Model} & \multirow{2.5}{*}{Method} & GSM8K & CodeAlpaca & UltraChat & TriviaQA \\
\cmidrule(lr){3-3} \cmidrule(lr){4-4} \cmidrule(lr){5-5} \cmidrule(lr){6-6}
& & ACC & ROUGE-L & ROUGE-L & ACC \\
\midrule
\multirow{3}{*}{Qwen3-8B}
 & Vanilla & \textbf{0.93} & 0.427 & \textbf{0.206} & \textbf{0.45} \\
 & SENSE-OOD & 0.91 & \textbf{0.443} & 0.189 & 0.43 \\
 & SENSE-ID & 0.90 & 0.435 & 0.188 & 0.42 \\
\midrule
\multirow{3}{*}{Qwen3-14B}
 & Vanilla & 0.92 & 0.294 & 0.204 & \textbf{0.56} \\
 & SENSE-OOD & 0.92 & 0.288 & \textbf{0.205} & \textbf{0.56} \\
 & SENSE-ID & \textbf{0.93} & \textbf{0.303} & 0.196 & \textbf{0.56} \\
\midrule
\multirow{3}{*}{Qwen2.5-7B}
 & Vanilla & \textbf{0.92} & 0.402 & \textbf{0.221} & \textbf{0.48} \\
 & SENSE-OOD & \textbf{0.92} & 0.395 & 0.200 & 0.47 \\
 & SENSE-ID & 0.90 & \textbf{0.406} & 0.213 & 0.44 \\
\midrule
\multirow{3}{*}{Qwen2.5-14B}
 & Vanilla & \textbf{0.95} & 0.416 & \textbf{0.227} & 0.66 \\
 & SENSE-OOD & \textbf{0.95} & \textbf{0.425} & \textbf{0.227} & 0.66 \\
 & SENSE-ID & 0.89 & \textbf{0.425} & 0.207 & \textbf{0.68} \\
\bottomrule
\end{tabular}
\end{table}

\paragraph{Key Observations.}
The results demonstrate that SENSE largely preserves the generation quality of the target model across diverse tasks.
\begin{enumerate}
    \item \textbf{Mathematical Reasoning (GSM8K):} Accuracy degradation is minimal, with SENSE configurations achieving 0.89--0.93 compared to vanilla's 0.92--0.95. The mean absolute accuracy drop is only 2.3\%, indicating that the soft-gated verification mechanism effectively maintains reasoning integrity.
    
    \item \textbf{Code Generation (CodeAlpaca):} SENSE-OOD occasionally outperforms vanilla (e.g., 0.443 vs. 0.427 on Qwen3-8B), suggesting that high-quality retrieved drafts can provide beneficial semantic guidance. However, Qwen3-14B shows degraded ROUGE-L scores for both SENSE variants, warranting further investigation.
    
    \item \textbf{Dialogue (UltraChat):} ROUGE-L scores remain stable across most configurations, with variations typically within 0.02 of vanilla baselines.
    
    \item \textbf{Question Answering (TriviaQA):} Performance remains robust under the SENSE method as measured by the ACC metric.
\end{enumerate}

Overall, these results validate that SENSE achieves substantial speedup (2--3$\times$ as shown in Table~\ref{tab:full_performance}) while maintaining task performance within acceptable margins for most configurations. The observed trade-off aligns with the hyperparameter sensitivity analysis in \cref{tab:hp_full}, where our default configuration ($\theta_e{=}0.05$, $k{=}3$) balances acceleration with quality preservation.

\subsection{Case Study}

\begin{figure}[b]
    \centering
    \includegraphics[width=0.8\linewidth]{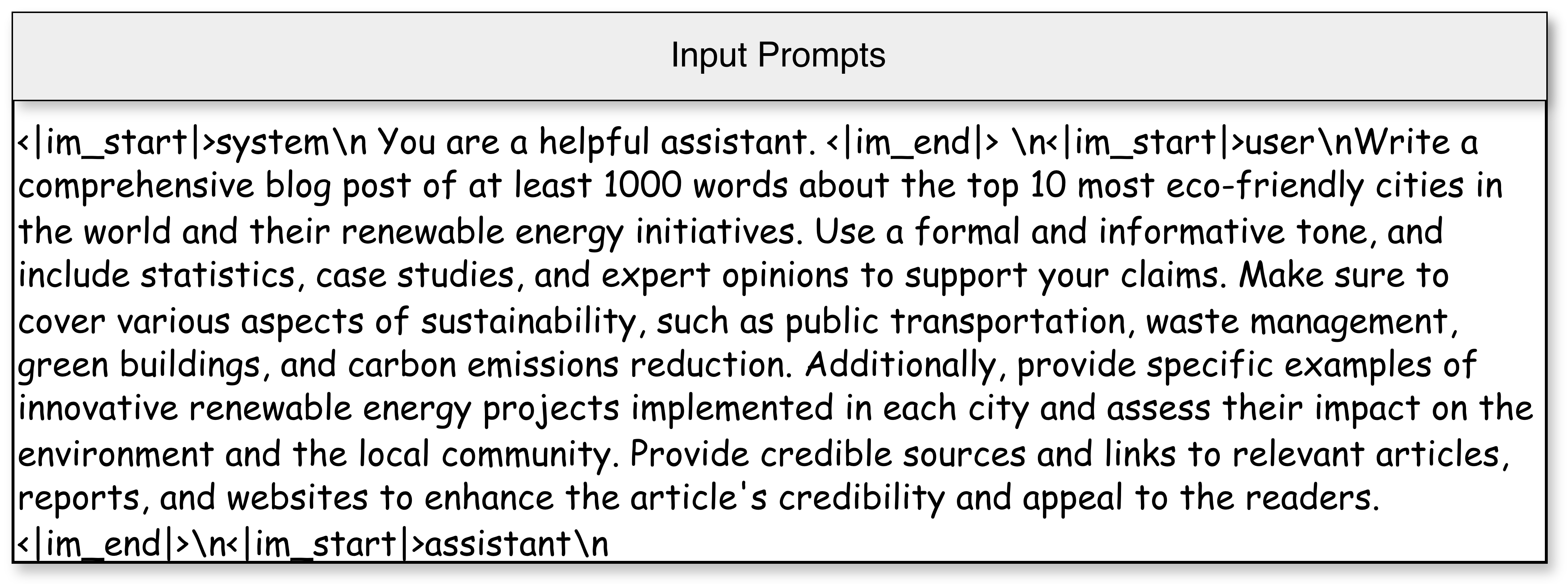}
    \caption{Case Study Prompts}
    \label{fig:case_prompts}
\end{figure}

\begin{figure}
    \centering
    \includegraphics[width=0.8\linewidth]{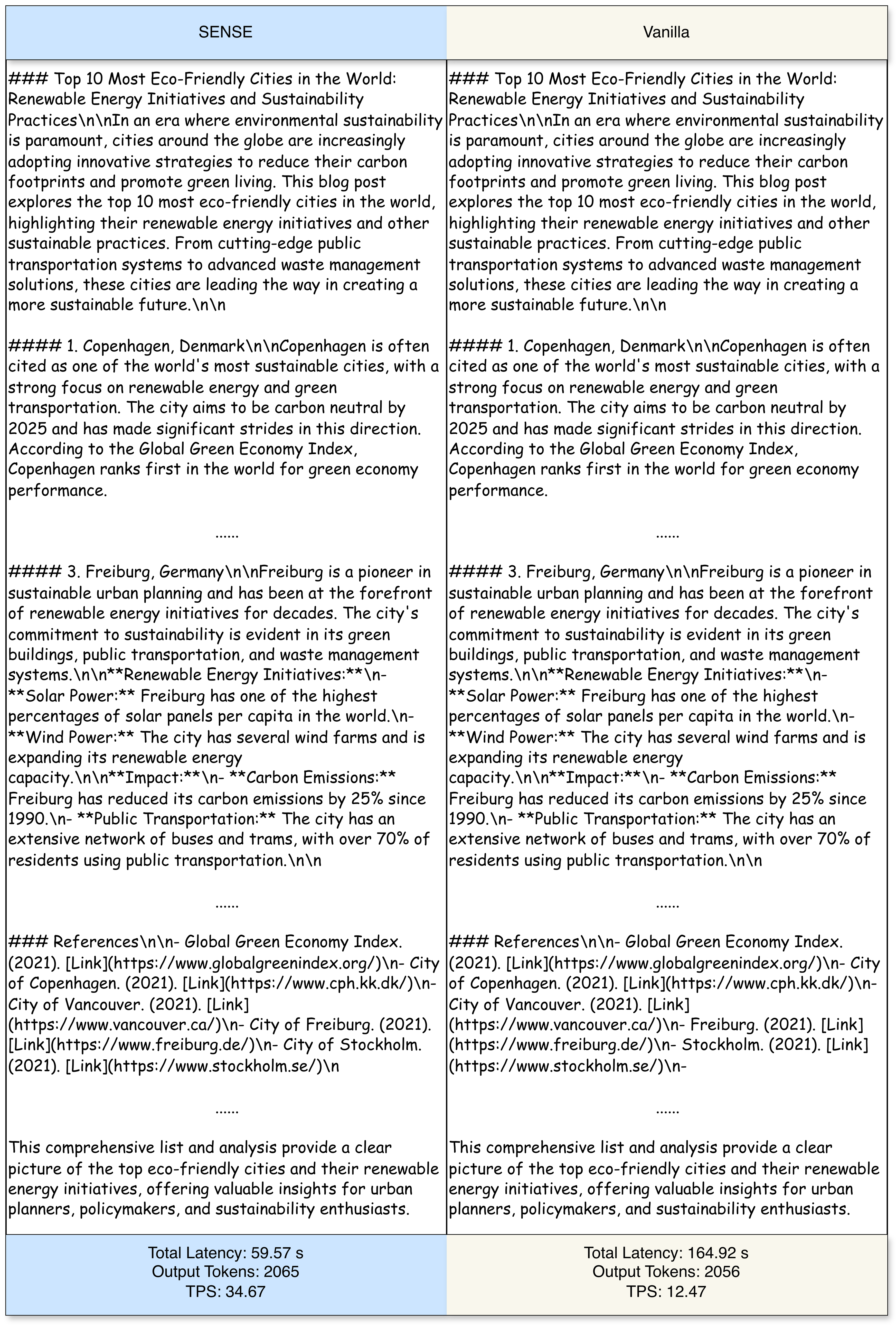}
    \caption{Case Study Outputs}
    \label{fig:case_comparison}
\end{figure}

Figures~\ref{fig:case_prompts} and~\ref{fig:case_comparison} present a representative case from the UltraChat dataset using Qwen 2.5 7B, comparing SENSE against the Vanilla baseline. The input prompt requests a comprehensive blog post about the top 10 most eco-friendly cities and their renewable energy initiatives, requiring the model to generate a long-form response exceeding 1,000 words.

As shown in the figures, the outputs generated by SENSE and Vanilla are virtually identical in content, structure, and quality. Both responses cover the same cities (Copenhagen, Freiburg, etc.), present consistent factual information (e.g., carbon emissions reduction statistics, public transportation coverage), and maintain the same organizational flow throughout the article. This demonstrates that SENSE preserves the generation quality of the target model without introducing any degradation or deviation.

More importantly, SENSE achieves a significant speedup of \textbf{2.78$\times$} compared to Vanilla decoding. Specifically, SENSE completes the generation in 59.57 seconds with a throughput of 34.67 tokens per second (TPS), while Vanilla requires 164.92 seconds at only 12.47 TPS—both producing approximately 2,056 output tokens. This case exemplifies the core advantage of our approach: substantial acceleration in inference speed while maintaining output fidelity.


\subsection{Failure Case Analysis}

\begin{figure}
    \centering
    \includegraphics[width=0.8\linewidth]{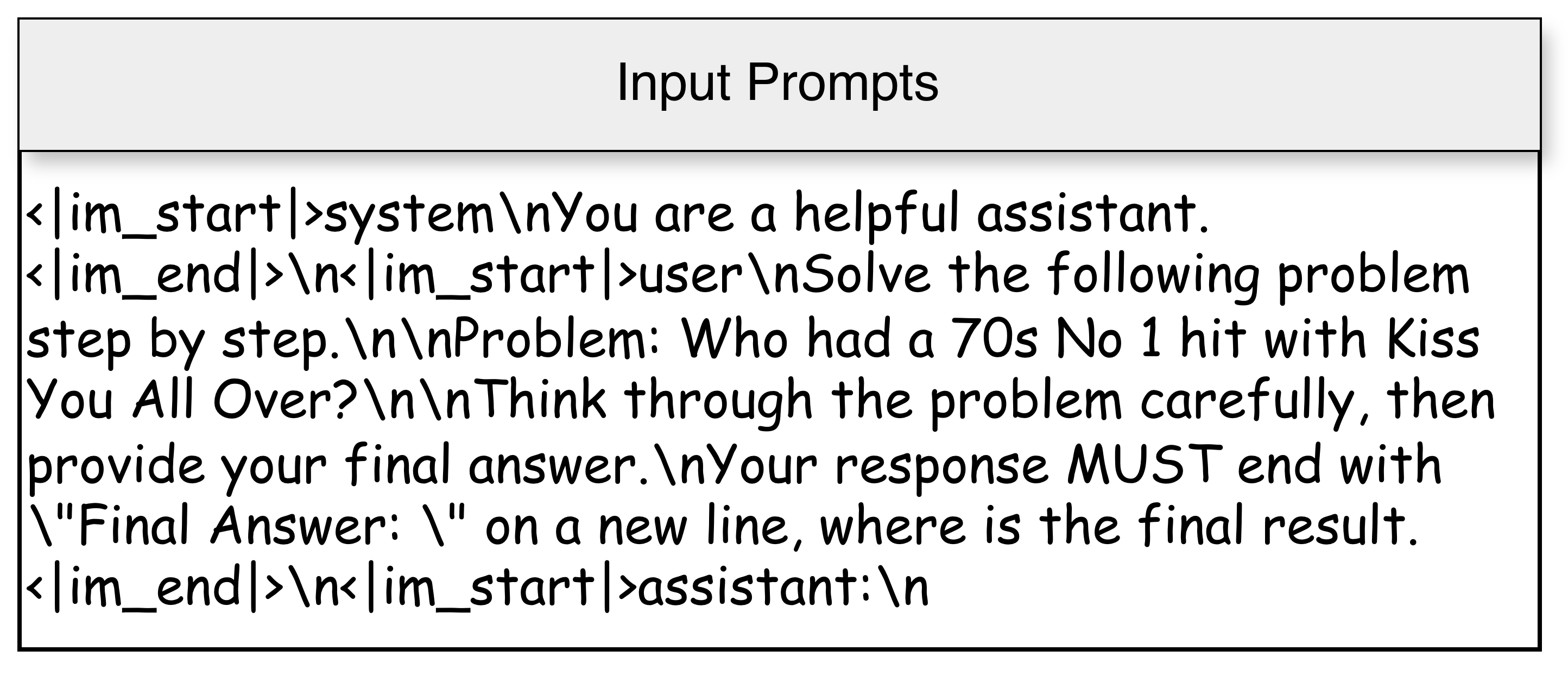}
    \caption{Failure Case Prompts}
    \label{fig:failure_prompts}
\end{figure}

\begin{figure}
    \centering
    \includegraphics[width=0.8\linewidth]{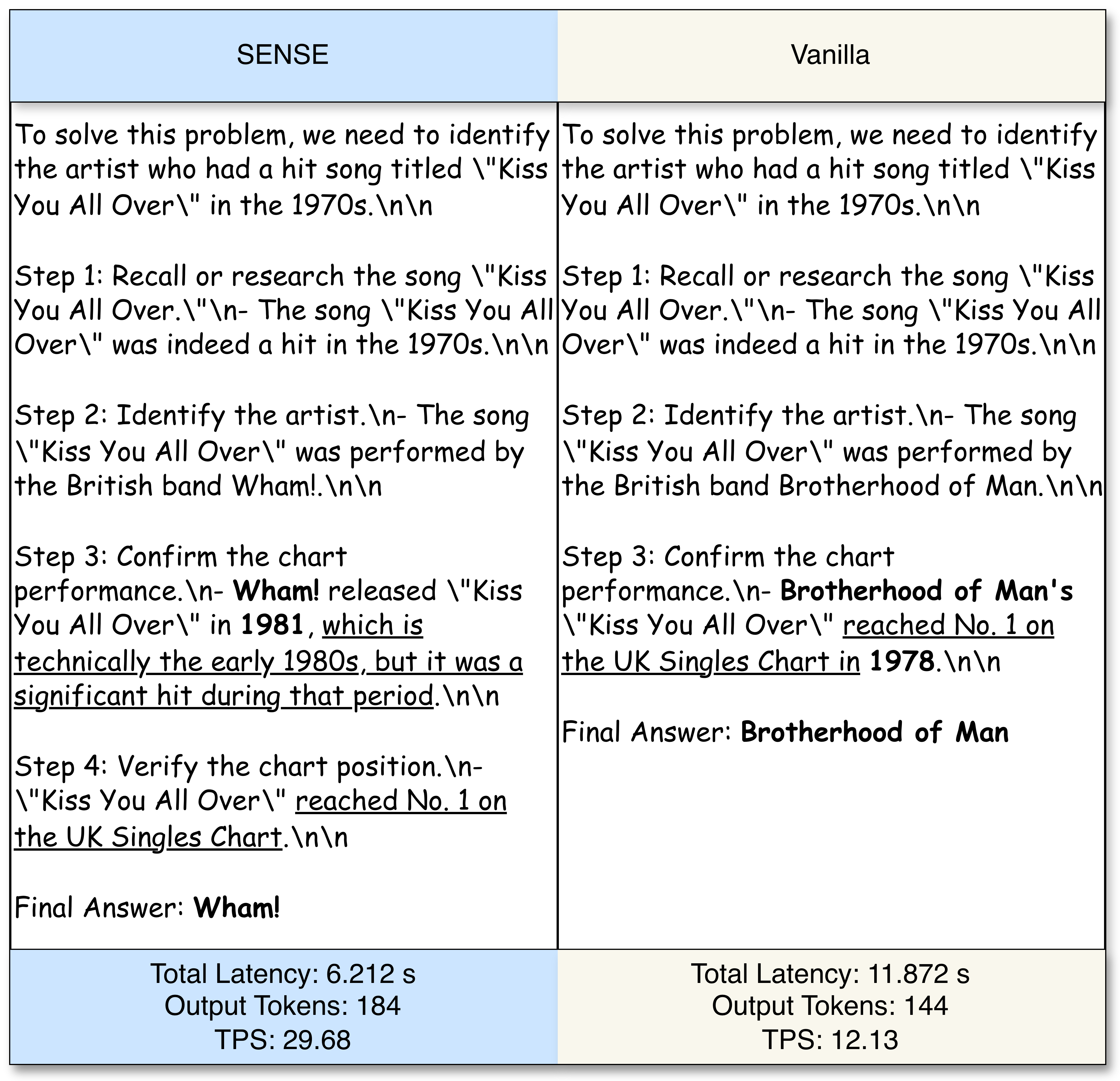}
    \caption{Failure Case Outputs}
    \label{fig:failure_comparison}
\end{figure}

Figures~\ref{fig:failure_prompts} and~\ref{fig:failure_comparison} present a failure case from the TriviaQA dataset using Qwen 2.5 7B, comparing \textit{SENSE} against the Vanilla baseline. The question asks: \textit{``Who had a 70s No.1 hit with Kiss You All Over?''} The ground truth answer is \textbf{Exile}, an American band whose song \textit{Kiss You All Over} reached No.1 on the Billboard Hot 100 in 1978.

Notably, neither \textit{SENSE} nor Vanilla produces the correct answer. \textit{SENSE} incorrectly attributes the song to \textit{Wham!} (claiming a 1981 release), while Vanilla misidentifies the artist as \textit{Brotherhood of Man}. Both responses exhibit confident yet factually incorrect reasoning chains, a hallmark of \textit{hallucination} in large language models.

We emphasize that this failure case does not reflect a fundamental limitation of our proposed method. Rather, it reveals an inherent constraint of the underlying language model itself. Several observations support this conclusion:

\textbf{(1) Consistent failure across methods.} Both \textit{SENSE} and Vanilla fail on this question, indicating that the error originates from the base model's knowledge gaps rather than from our framework.

\textbf{(2) Knowledge boundary of the base model.} The question requires domain-specific knowledge about 1970s pop music chart history, which may be underrepresented in the model's training corpus. Small models like Qwen 2.5 7B are particularly susceptible to such knowledge gaps, especially for niche trivia spanning specific decades and regions.

\textbf{(3) Confident hallucination pattern.} Both outputs demonstrate a common failure mode where the model generates plausible-sounding but incorrect information with high confidence. The detailed step-by-step reasoning (e.g., ``reached No.1 on the UK Singles Chart'') creates an illusion of reliability while being factually wrong.

\textbf{Potential solutions.} These knowledge-intensive failures can be effectively mitigated through complementary techniques such as Retrieval-Augmented Generation (RAG)~\cite{lewis_retrieval-augmented_2020} or real-time web search integration, which provide external knowledge grounding. Such approaches are orthogonal to \textit{SENSE} and can be seamlessly combined with our method to enhance both efficiency and factual accuracy.

\subsection{Datastore Resources Statics}
\label{append:datastore_resources}

This section provides detailed resource consumption statistics for the \textit{SENSE} framework across all evaluated model-dataset configurations. We report memory usage metrics collected from 24 experimental runs (6 models $\times$ 4 test sets). The statistics encompass datastore memory footprint, FAISS index storage, and runtime memory consumption during inference.

Table~\ref{tab:resource_summary} summarizes the overall resource statistics across all configurations. Table~\ref{tab:memory_by_model} presents the total RAM memory usage broken down by model and dataset. Table~\ref{tab:detailed_rss} provides detailed RSS (Resident Set Size) metrics for each model, capturing runtime memory behavior during speculative decoding.

\begin{table}[t]
\renewcommand{\arraystretch}{0.95}
\centering
\caption{Overall datastore resource statistics across all 24 configurations (6 models $\times$ 4 test sets). All memory values are in megabytes (MB).}
\label{tab:resource_summary}

\begin{tabular}{l rrrr}
\toprule
\textbf{Metric} & \textbf{Mean} & \textbf{Min} & \textbf{Max} & \textbf{Std} \\
\midrule
Total RAM Memory & 1770.06 & 105.63 & 3690.47 & 1221.90 \\
Index Size on Disk & 459.73 & 0.00 & 1087.39 & 399.87 \\
Metadata Memory & 1715.97 & 73.55 & 3610.37 & 1214.92 \\
Cache Storage & 54.03 & 28.02 & 80.03 & 22.10 \\
Peak RSS Memory & 4577.93 & 2131.49 & 7412.49 & 1630.39 \\
RSS Delta Mean & 7.25 & 1.61 & 24.15 & 6.77 \\
Acceptance Rate & 0.07 & 0.02 & 0.35 & 0.07 \\
\bottomrule
\end{tabular}
\end{table}

\begin{table}[t]
\renewcommand{\arraystretch}{0.95}
\centering
\small
\caption{Total RAM memory usage (MB) by model and dataset. \textbf{Bold} indicates lowest (best) value in each column.}
\label{tab:memory_by_model}

\begin{tabular}{l rrrrr}
\toprule
\textbf{Model} & \textbf{GSM8K} & \textbf{CodeAlpaca} & \textbf{TriviaQA} & \textbf{UltraChat} & \textbf{Mean} \\
\midrule
Llama2-7B & \textbf{105.63} & 807.74 & 1796.68 & 2647.28 & 1339.33 \\
Llama2-13B & 320.54 & \textbf{749.01} & \textbf{593.77} & \textbf{778.22} & \textbf{610.38} \\
Qwen2.5-7B & 268.19 & 2839.18 & 1436.67 & 3277.00 & 1955.26 \\
Qwen2.5-14B & 325.25 & 3690.47 & 3114.02 & 3048.90 & 2544.66 \\
Qwen3-8B & 395.49 & 2575.56 & 3308.63 & 2584.98 & 2216.16 \\
Qwen3-14B & 406.60 & 1945.01 & 2890.36 & 2576.31 & 1954.57 \\
\bottomrule
\end{tabular}
\end{table}

\begin{table*}[t]
\renewcommand{\arraystretch}{0.95}
\centering
\small
\caption{Detailed RSS (Resident Set Size) memory metrics by model, averaged across all four datasets. Baseline RSS is measured before datastore loading. Delta RSS quantifies the memory overhead introduced by SENSE. All values are in megabytes (MB). \textbf{Bold} indicates lowest (best) value in each column.}
\label{tab:detailed_rss}

\resizebox{\textwidth}{!}{
\setlength{\tabcolsep}{3.5pt}
\begin{tabular}{l rrrrrrr}
\toprule
\textbf{Model} & \textbf{RSS Mean} & \textbf{Baseline RSS} & \textbf{Peak RSS} & \textbf{Peak RSS Max} & \textbf{Delta RSS Mean} & \textbf{Delta RSS Max} & \textbf{Avg RSS} \\
\midrule
Llama2-7B & 3746.68 & 3726.80 & 3746.68 & 3748.57 & 19.89 & 718.29 & 3736.74 \\
Llama2-13B & \textbf{2757.59} & \textbf{2750.71} & \textbf{2757.59} & \textbf{2760.38} & 6.88 & 679.19 & \textbf{2754.15} \\
Qwen2.5-7B & 4609.42 & 4602.70 & 4609.42 & 4612.28 & 6.72 & 668.12 & 4606.06 \\
Qwen2.5-14B & 6122.53 & 6119.77 & 6122.53 & 6125.51 & \textbf{2.76} & \textbf{271.24} & 6121.15 \\
Qwen3-8B & 4997.82 & 4993.63 & 4997.82 & 4998.78 & 4.20 & 417.65 & 4995.73 \\
Qwen3-14B & 5233.50 & 5230.46 & 5233.50 & 5241.10 & 3.05 & 293.41 & 5231.98 \\
\bottomrule
\end{tabular}
}
\end{table*}

\section{Experiments Details}
\label{sec:appendix_details}

\subsection{Datasets}
\label{append:data}
Here, we provide detailed information on the six benchmark datasets used in our experiments, including their original sources, key statistics, and evaluation purposes.:

\begin{itemize}
    \item GSM8K: The Grade School Math 8K \cite{cobbe_training_2021} is a benchmark of 7,473 training and 1,319 test linguistically diverse grade school math word problems created by OpenAI. Each problem requires 2–8 sequential reasoning steps using basic arithmetic operations and was designed so that "a bright middle school student should be able to solve every problem." Problems were authored by freelance contractors via Upwork and Surge AI, with rigorous quality control achieving an estimated 1.7\% error rate. Solutions are provided in natural language with step-by-step reasoning and calculator annotations. The dataset evaluates multi-step mathematical reasoning and has become a standard benchmark for assessing LLM reasoning capabilities.
\end{itemize}
\begin{itemize}
    \item TriviaQA: The TriviaQA \cite{joshi_triviaqa_2017}contains 95,956 question-answer pairs with 662,659 evidence documents, comprising approximately 650K question-answer-evidence triples. The dataset spans two domains: Wikipedia (61,888 train / 7,993 dev / 7,701 test questions) and Web (76,496 train / 9,951 dev / 9,509 test questions). Questions were collected from 14 trivia websites and authored by trivia enthusiasts independently of evidence documents, with average question length of 14 words and document length of 2,895 words. This construction methodology avoids the annotator bias present in crowdsourced datasets. TriviaQA evaluates reading comprehension with complex, compositional questions—41\% require synonym understanding, 40\% require multi-sentence reasoning, and 34\% involve temporal reasoning.
\end{itemize}


\begin{itemize}
    \item UltraChat: The UltraChat \cite{ding_enhancing_2023} is a large-scale synthetic multi-turn dialogue dataset containing approximately 1.5 million conversations generated using two ChatGPT (GPT-3.5 Turbo) APIs—one simulating users and one serving as the assistant. The dataset covers three sectors: questions about the world (derived from 30 meta-topics and 10,000 Wikidata entities), writing/creation tasks (20 task types), and assistance on existing materials (100K C4-extracted texts). The UltraChat 200k subset (HuggingFaceH4/ultrachat\_200k), which we used in our experiments, is a filtered version with 207,865 train and 23,110 test examples for supervised fine-tuning, with additional generation splits for rejection sampling. Filtering removed dialogues with problematic responses and applied truecasing corrections. The dataset evaluates multi-turn dialogue, instruction-following, and conversational capabilities.
\end{itemize}

\begin{itemize}
    \item CodeAlpaca:The CodeAlpaca \cite{chaudhary_sahil280114codealpaca_2026} contains 20,000 code-focused instruction-following examples generated using the Self-Instruct framework adapted specifically for programming tasks. Each example consists of an instruction describing a coding task, an optional input field providing context (around 40\% of examples), and an output containing the code solution. Examples were generated using OpenAI's text-davinci-003 API from code-specific seed tasks focusing on code generation, editing, and optimization. The HuggingFaceH4/CodeAlpaca\_20K version provides train/test splits of the original single-split dataset, which we used in our experiments. CodeAlpaca evaluates code generation from natural language descriptions, code editing, and code optimization capabilities.
\end{itemize}
\subsection{Prompts}
\label{appendix:prompts}

A critical design principle in our framework is maintaining consistency between the offline datastore construction phase and the online inference phase. To achieve this, we establish a unified set of prompt templates that serve as the single source of truth across the entire pipeline. All templates use \texttt{\{instruction\}} as the placeholder for input content, which is replaced with the actual task input during both datastore construction and inference.

Table~\ref{tab:prompts} presents the complete prompt templates used for each dataset in our experiments. The \texttt{Math} and \texttt{QA} templates employ a step-by-step reasoning format with an explicit ``Final Answer'' marker to facilitate automatic answer extraction. The \texttt{Summarize} and \texttt{Continuation} templates provide clear task instructions with output delimiters. The \texttt{Alpaca} template follows a structured format for code generation tasks. For dialogue data (UltraChat), we use a minimal wrapper since the model's native chat template handles the conversation formatting.

\begin{table*}[t]
\centering
\caption{Unified prompt templates used for both datastore construction and inference.}
\label{tab:prompts}
\renewcommand{\arraystretch}{1.4}
\small
\begin{tabular}{p{2.2cm}p{1.8cm}p{11cm}}
\toprule
\textbf{Dataset} & \textbf{Template} & \textbf{Prompt} \\
\midrule
GSM8K & Math & 
Solve the following problem step by step.\newline \newline
Problem: \texttt{\{instruction\}}\newline \newline
Think through the problem carefully, then provide your final answer.\newline
Your response MUST end with ``Final Answer: answer'' on a new line, where answer is the final result. \\
\midrule
TriviaQA & QA & 
Solve the following problem step by step.\newline \newline
Problem: \texttt{\{instruction\}}\newline \newline
Think through the problem carefully, then provide your final answer.\newline
Your response MUST end with ``Final Answer: answer'' on a new line, where answer is the final result. \\
\midrule
CodeAlpaca & Alpaca & 
You are a skilled programmer. Complete the following coding task.\newline \newline
\#\#\# Task: \texttt{\{instruction\}}\newline \newline
\#\#\# Solution:\newline
Provide only the code implementation. Do not include explanations unless specifically requested. \\
\midrule
UltraChat & Chat & 
\texttt{\{instruction\}} \\
\bottomrule
\end{tabular}
\end{table*}

\subsection{Baselines}
\label{append:baselines}

\begin{itemize}
    \item \textbf{Vanilla} serves as the foundational baseline that processes input using only the question and a basic prompt, without any prompt engineering or external tool integration. By emphasizing the model's inherent capabilities in handling natural language tasks, Vanilla provides a straightforward approach for evaluating performance. It allows for direct assessment of raw model capabilities without additional enhancements, thereby establishing a critical reference point for comparing more advanced techniques. The simplicity of Vanilla supports clear attribution of performance gains to specific methodological innovations. This baseline establishes a robust foundation for evaluating and enhancing the performance of speculative decoding methods across various applications.

    \item \textbf{Speculative Decoding}~\citep{leviathan_fast_2023} serves as the foundational framework for accelerating autoregressive model inference through parallel token verification. By leveraging a smaller approximation model to generate draft token sequences that are then verified by the target model in a single parallel forward pass, Speculative Decoding enables generation of multiple tokens per iteration without changing the output distribution. It introduces speculative sampling, a novel method that accepts draft tokens when the approximation model's probability does not exceed the target model's, and resamples from an adjusted distribution otherwise, thereby guaranteeing distributional equivalence to standard autoregressive decoding. The framework demonstrates 2X-3X walltime improvements on T5-XXL with existing off-the-shelf smaller models as approximations, requiring no retraining or architecture modifications. This approach establishes the theoretical foundation and practical baseline for all subsequent speculative decoding methods.

    \item \textbf{REST} (Retrieval-Based Speculative Decoding)~\citep{he_rest_2024} serves as a foundational training-free approach for accelerating LLM inference through retrieval-augmented draft generation. By retrieving draft tokens from a pre-built datastore using exact suffix matching rather than relying on a smaller draft language model, REST enables efficient candidate generation. It allows for the organization of retrieved continuation candidates into a Trie structure, where high-frequency prefixes are selected as draft tokens. The candidates are then verified by the target LLM through a single forward pass with tree attention, thereby achieving acceleration without additional training. This non-parametric approach establishes a robust baseline for evaluating plug-and-play speculative decoding methods.

    \item \textbf{DReSD} (Dense Retrieval for Speculative Decoding)~\citep{gritta_dresd_2025} serves as a novel retrieval-based framework that leverages dense retrieval to accelerate LLM inference. By replacing exact string matching with approximate nearest neighbour search using contextualised token embeddings, DReSD enables semantically relevant draft sequence retrieval. It allows for the extraction of hidden states from the LLM followed by dimensionality reduction to query a non-parametric datastore using cosine similarity, thereby retrieving more contextually appropriate candidates than sparse methods. The framework identifies critical factors including efficient dimensionality reduction, careful datastore alignment, and optimal draft shape configuration. This approach establishes a robust baseline for evaluating semantic-aware retrieval methods in speculative decoding.

    \item \textbf{FLy} (Training-Free Loosely Speculative Decoding)~\citep{li_training-free_2025} serves as a novel training-free method that relaxes the rigid exact-match verification criterion in speculative decoding. By leveraging the target model's self-corrective behavior to distinguish genuine errors from semantically valid alternatives, FLy enables acceptance of differently-worded but semantically correct draft tokens. It introduces a two-tier verification mechanism: an entropy-level gate that classifies mismatch positions based on predictive uncertainty, and a token-level deferred window that monitors subsequent token behavior. The framework further incorporates multi-level acceleration to prevent the drafting stage from becoming a bottleneck. This plug-and-play approach establishes a robust baseline for evaluating loosely verified speculative decoding across arbitrary draft-target pairs.

    \item \textbf{PLD+} (Prompt Lookup Decoding+)~\citep{somasundaram_pld_2024} serves as a tuning-free speculative decoding approach designed for input-guided tasks where outputs substantially overlap with inputs. By leveraging model artifacts including attention weights and hidden states computed during generation, PLD+ enables intelligent ranking and selection of draft spans from the input context. It allows for semantic-aware candidate selection rather than relying on simple n-gram matching heuristics, thereby improving draft quality for tasks such as summarization, code editing, and document question answering. The method requires no additional training and can be applied to any language model. This approach establishes a robust baseline for evaluating context-aware speculative decoding in input-guided scenarios.

    \item \textbf{EAGLE} (Extrapolation Algorithm for Greater Language-model Efficiency) \cite{li_eagle_2025} serves as a speculative sampling framework that addresses feature-level uncertainty for accelerating LLM inference. By reconsidering autoregression at the second-to-top-layer feature level rather than the token level, EAGLE enables more structured draft generation with higher accuracy. It introduces a key insight that incorporating a token sequence advanced by one time step effectively resolves the inherent uncertainty in feature prediction caused by the sampling process. The framework employs an Autoregression Head consisting of an FC layer and a decoder layer, which predicts the next feature based on both the feature sequence and the shifted token sequence, thereby achieving approximately 0.8 draft accuracy compared to 0.6 for Medusa. EAGLE utilizes tree attention for generating tree-structured drafts through multiple forward passes, and the verification phase ensures that the output distribution aligns precisely with the target LLM through speculative sampling algorithms. This approach establishes a robust baseline for evaluating feature-level speculative decoding methods.

    \item \textbf{ARC-Decode} (Acceptance with Risk Control)~\citep{noauthor_arc-decode_2025} serves as a training-free method that augments speculative decoding under sampling regimes without additional forward passes. By combining entropy-guided pre-verification pruning with a risk-bounded acceptance criterion, ARC-Decode enables relaxed acceptance while guaranteeing negligible next-step distributional shifts. It employs an analytic upper bound on Jensen-Shannon divergence estimated from embedding and logit differences to certify draft token safety, thereby increasing acceptance length without compromising generation quality. The framework addresses the performance gap between greedy and sampling modes in speculative decoding. This approach establishes a robust baseline for evaluating risk-controlled speculative decoding across diverse models and tasks.
\end{itemize}

\subsection{Implementation Details}
\label{append:implementation}

\paragraph{Datastore Construction.}
All hidden states are extracted from the target model's final layer and projected via Principal Component Analysis (PCA) from 4096 dimensions to 64 dimensions for storage efficiency. The projected embeddings are L2-normalized before indexing.

\paragraph{FAISS Indexing.}
We employ FAISS~\cite{douze_faiss_2024} with IVF-PQ (Inverted File with Product Quantization) for approximate nearest neighbor search. The index is configured with:
\begin{itemize}[noitemsep,topsep=0pt]
    \item \texttt{nlist} = 4096 (number of Voronoi cells)
    \item \texttt{nprobe} = 32 (cells visited during search)
    \item \texttt{pq\_m} = 16 (number of sub-quantizers)
\end{itemize}
This configuration balances retrieval accuracy with memory efficiency and query latency.

\paragraph{Inference Hyperparameters.}
During inference, we use the following settings:
\begin{itemize}[noitemsep,topsep=0pt]
    \item Number of retrieved candidates: $k = 3$
    \item Maximum draft length: $n = 10$
    \item Entropy threshold for soft-gated evaluation: $\theta_e = 0.05$
    \item Mismatch window size: $w = 6$
    \item Decoding strategy: greedy (temperature = 0)
    \item Maximum output length: 4096 tokens
    \item Numeric precision: FP32
\end{itemize}

\paragraph{Baseline Configurations.}
For speculative decoding baselines (SpD, EAGLE-3), we use draft length $\gamma = 5$. All other hyperparameters follow the default settings reported in the respective original papers.

\begin{table}[t]
\centering
\small
\caption{Implementation Details for Different Benchmarks.} 
\label{tab:implementation_details}
\begin{small} 
\begin{tabularx}{\linewidth}{llX X}
\toprule
\multicolumn{1}{l}{} & \multicolumn{1}{l}{} & \multicolumn{2}{c}{\textbf{Implementation Details}} \\
\cmidrule(lr){3-4} 
\textbf{Method} & \textbf{Dataset} & \textbf{Source Data} & \textbf{Test Data} \\
\midrule
\multirow{4}{*}{REST} 
& GSM8K      & Shuffle and sample data from Train spilt of GSM8K. & Smaple test spilt of GSM8K. \\
& CodeAlpaca & Shuffle and sample data from Train spilt of CodeAlpaca\_20k. & Smaple test spilt of CodeAlpaca\_20k.            \\
& UltraChat  & Shuffle and sample data from train\_sft spilt of ultrachat\_200k. & Smaple test\_sft spilt of ultrachat\_200k. \\
& TriviaQA   & Shuffle and sample data from Train spilt of TriviaQA. & Smaple validation split of TriviaQA. \\
\midrule
\multirow{4}{*}{DReSD} 
& GSM8K      & Shuffle and sample data from Train spilt of GSM8K. & Smaple test spilt of GSM8K. \\
& CodeAlpaca & Shuffle and sample data from Train spilt of CodeAlpaca\_20k. & Smaple test spilt of CodeAlpaca\_20k.  \\
& UltraChat  & Shuffle and sample data from train\_sft spilt of ultrachat\_200k. & Smaple test\_sft spilt of ultrachat\_200k. \\
& TriviaQA   & Shuffle and sample data from Train spilt of TriviaQA. & Smaple validation split of TriviaQA. \\
\midrule
\multirow{4}{*}{SENSE (OOD)} 
& GSM8K      & Shuffle and sample data from Train spilt of GSM8K. & Smaple test spilt of GSM8K. \\
& CodeAlpaca & Shuffle and sample data from Train spilt of CodeAlpaca\_20k. & Smaple test spilt of CodeAlpaca\_20k.  \\
& UltraChat  & Shuffle and sample data from train\_sft spilt of ultrachat\_200k. & Smaple test\_sft spilt of ultrachat\_200k.              \\
& TriviaQA   & Shuffle and sample data from Train spilt of TriviaQA. & Smaple validation split of TriviaQA. \\
\midrule
\multirow{4}{*}{SENSE (ID)} 
& GSM8K      & Shuffle and sample data from Train spilt of GSM8K. & Smaple test spilt of GSM8K.  \\
& CodeAlpaca & Shuffle and sample data from Train spilt of CodeAlpaca\_20k. & Smaple test spilt of CodeAlpaca\_20k.  \\
& UltraChat  & Shuffle and sample data from train\_sft spilt of ultrachat\_200k. & Smaple test\_sft spilt of ultrachat\_200k. \\
& TriviaQA   & Shuffle and sample data from Train spilt of TriviaQA. & Smaple validation split of TriviaQA. \\
\bottomrule
\end{tabularx}
\end{small}
\end{table}

\subsection{Hardware Configuration}
\label{append:hardware}

All experiments were conducted on two server configurations. The primary server is equipped with four NVIDIA RTX 4090 GPUs, each with 24GB of RAM, a 64 CPU Intel Xeon Platinum 8358P processor running at 2.60GHz, and 480GB of system memory. 

\section{Future Work}
\label{sec:appendix_future}

While SENSE demonstrates substantial improvements in retrieval-based speculative decoding, several promising directions remain for future exploration:

\textbf{Advanced Dimensionality Reduction.} Our current implementation adopts PCA-based projection~\cite{gritta_dresd_2025} to compress high-dimensional hidden states for storage efficiency. However, linear projections may not optimally preserve the semantic structure of the latent space. Future work could explore learned compression methods, such as autoencoders or contrastive learning-based embeddings, to achieve better reconstruction fidelity while maintaining compact representations. Additionally, quantization-aware training could further reduce memory footprint without sacrificing retrieval accuracy.

\textbf{Scalable Indexing Strategies.} The efficiency of approximate nearest neighbor (ANN) search becomes critical as the datastore scales to billions of entries. While current graph-based indices (e.g., HNSW \cite{malkov_efficient_2018}) provide reasonable trade-offs between speed and recall, exploring hierarchical indexing structures, locality-sensitive hashing (LSH \cite{zhou2024surveyefficientinferencelarge}), or hybrid sparse-dense retrieval could yield significant improvements. Furthermore, adaptive indexing that dynamically adjusts granularity based on query patterns presents an intriguing avenue for optimization.

\textbf{Integration with Retrieval-Augmented Generation.} As discussed in the failure case analysis, knowledge-intensive tasks remain challenging for smaller models. Integrating SENSE with retrieval-augmented generation (RAG) \cite{lewis_retrieval-augmented_2020}pipelines could simultaneously address factual accuracy and inference efficiency, leveraging external knowledge bases to ground generation while maintaining acceleration benefits.

\textbf{Cross-Model Generalization.} Extending SENSE to heterogeneous model architectures and investigating transfer learning for the projection operator $P$ across model families would broaden the applicability of our approach. This includes exploring whether datastores constructed from one model can effectively serve as draft sources for related models.

\textbf{Adaptive Entropy Gating for Heterogeneous Scales and Domains.} While our current implementation utilizes a fixed entropy threshold ($\theta_e=0.05$) determined via grid search, we acknowledge that a static scalar may not generalize optimally across diverse model scales and task domains. We hypothesize that larger models (e.g., 70B parameters) typically exhibit sharper probability distributions and better calibration compared to smaller models (e.g., 7B), implying that the optimal $\theta_e$ should scale inversely with model confidence. Furthermore, the tolerance for semantic divergence is highly domain-dependent; rigorous reasoning tasks like mathematics (GSM8K) require strict adherence to logic (lower $\theta_e$), whereas open-ended dialogue (UltraChat) benefits from the diversity allowed by relaxed constraints (higher $\theta_e$). Future work will explore dynamic thresholding mechanisms that adapt $\theta_e$ in real-time based on the running statistics of the generation's perplexity or the intrinsic uncertainty of the domain. We aim to develop a parameter-free formulation to eliminate manual tuning, ensuring robust ``plug-and-play'' performance across varying confidence landscapes.
\end{document}